\documentclass[letterpaper,10pt]{article}

\usepackage[margin=1.25in]{geometry}
\usepackage[round]{natbib}
\usepackage{afterpage}
\usepackage{pratik_icml}

\makeatletter
\renewcommand\AB@affilsepx{, \protect\Affilfont}

\renewcommand\Affilfont{\fontsize{9.5}{14.4}\selectfont}
\makeatother

\title{A Picture of the Space of Typical Learnable Tasks }

\author[1]{Rahul Ramesh}
\author[1]{Jialin Mao}
\author[2]{Itay Griniasty}
\author[1]{Rubing Yang}
\author[2]{Han Kheng Teoh}
\author[3]{\\\vspace*{-0.07cm}Mark K. Transtrum}
\author[2]{James P. Sethna}
\author[1]{Pratik Chaudhari\vspace*{-0.2cm}}

\affil[1]{University of Pennsylvania}
\affil[2]{Cornell University}
\affil[3]{Brigham Young University}

\date{\vspace*{-0.6cm}\small Email: \texttt{rahulram,pratikac@seas.upenn.edu; 
jmao,rubingy@sas.upenn.edu; \protect\\
ig324,ht452,sethna@cornell.edu; mkt24@byu.edu
}}

\def \yvec {{\vec{y}}}
\def \yvecs {{\vec{y^*}}}

\def \Ps {P_*}
\def \Pz {P_0}

\def \d {\text{d}}
\def \dB {\text{d}_{\text{B}}}
\def \dG {\text{d}_{\text{G}}}

\def \dtraj {\text{d}_{\text{traj}}}

\def \tt {{\tilde{\t}}}

\def \PU {P^\text{U}}
\def \tot {\t^{1 \to 2}}
\def \tto {\t^{2 \to 1}}
\def \tou {\t^{1 \to \text{U}}}
\def \ttu {\t^{2 \to \text{U}}}

\def \wb {{w_1}}
\def \wc {{w_2}}

\graphicspath{{./fig/}{../}}
\usepackage[normalem]{ulem}

\begin{document}

\maketitle
\let\thefootnote\relax\footnotetext{Proceedings of the 40$^{\text{th}}$ International Conference on Machine Learning, Honolulu, Hawaii, USA. Copyright 2023 by the authors. \vspace{0.2em}}

\let\thefootnote\relax\footnotetext{We have developed the techniques discussed here in more detail in~\citet{mao2023training}. It shows that that the training process explores an effectively low-dimensional manifold. Networks with a wide range of architectures, sizes, trained using different optimization methods, regularization techniques, data augmentation techniques, and weight initializations lie on the same manifold in the prediction space}

\begin{abstract}
We develop information geometric techniques to understand the representations learned by deep networks when they are trained on different tasks using supervised, meta-, semi-supervised and contrastive learning. We shed light on the following phenomena that relate to the structure of the space of tasks:
\begin{enumerate*}[(1)]
\item the manifold of probabilistic models trained on different tasks using different representation learning methods is effectively low-dimensional;
\item supervised learning on one task results in a surprising amount of progress even on seemingly dissimilar tasks; progress on other tasks is larger if the training task has diverse classes;
\item the structure of the space of tasks indicated by our analysis is consistent with parts of the Wordnet phylogenetic tree;
\item episodic meta-learning algorithms and supervised learning traverse different trajectories during training but they fit similar models eventually;
\item contrastive and semi-supervised learning methods traverse trajectories similar to those of supervised learning.
\end{enumerate*}
We use classification tasks constructed from the CIFAR-10 and Imagenet datasets to study these phenomena. Code is available at \href{https://github.com/grasp-lyrl/picture_of_space_of_tasks}{https://github.com/grasp-lyrl/picture\_of\_space\_of\_tasks}.
\end{abstract}


\section{Introduction}
\label{s:intro}

Exploiting data from related tasks to reduce the sample complexity of learning a desired task, is an idea that lies at the heart of burgeoning fields like transfer, multi-task, meta, few-shot, semi- and self-supervised learning. These algorithms, aided by the ability of deep networks to learn flexible representations, have shown an impressive ability to predict well on new tasks. These algorithms are very different from each other but it stands to reason they must be exploiting some structure in the space of tasks. We do not yet know what the structure in the space of tasks is precisely (see~\cref{s:related} for a discussion of related work). The goal of this paper is to characterize this structure and shed light on \emph{why} these existing algorithms are successful.

We develop information-geometric techniques to analyze representations learned by different algorithms on different tasks. The key idea of this paper is to think of a deep network with weights $w$ trained on a task as a probabilistic model
\[
    P_w(\yvec) = \prod_{n=1}^N p_w^n(y_n)
\]
where $\yvec = (y_1, \ldots, y_N)$ denotes any sequence of outputs (each $y_n \in \cbr{1,\ldots,C}$ classes) on $N$ independent and identically distributed samples and $p_w^n(y_n)$ denotes the probability that sample $x_n$ belongs to class $y_n$ as predicted by a deep network with weights $w$. We instantiate the technical machinery of information geometry using this $NC$-dimensional object to study different probabilistic models fitted to the task irrespective of which representation learning algorithm, e.g., supervised learning, meta-learning, etc., or what neural architecture was used to fit the probabilistic model. This construction circumvents the enormous diversity of algorithms, architectures with different feature spaces and training methods across these different sub-fields and provides us with a single space to study these models in --- the prediction space of the model.

\subsection{Contributions}

We discuss theoretical and computational tools to study such probabilistic models in~\cref{s:methods}. Many of these tools are developed in more detail in~\citet{mao2023training}. These tools are used to visualize these very high-dimensional objects, to compute geodesics on such manifolds, to interpolate checkpoints along training trajectories into continuous curves, and to map models trained on different tasks into a unique prediction space. We point these technical tools to understanding the structure of the space of learnable tasks and study different representation algorithms such as supervised, transfer, meta, semi- and self-supervised learning. We report the following findings in~\cref{s:results}:

\begin{enumerate}[(1),nosep]
\item The manifold of probabilistic models trained on different tasks using different representation learning methods is effectively low-dimensional. This dimensionality is very small: for Imagenet where our probabilistic models are in $10^7$ dimensions, the top 3 dimensions preserve 80.02\% of the pairwise distances between 2430 models trained on different sub-tasks of Imagenet.
\item Supervised learning on one task results in a surprising amount of progress on seemingly dissimilar tasks (informally, ``progress'' means that the representation learned on one can be used to make accurate predictions on other tasks; this is defined in~\cref{eq:tw}); progress on other tasks is larger if the training task has diverse classes.
\item Structure of the space of tasks indicated by our analysis is consistent with parts of the Wordnet phylogenetic tree.
\item Episodic meta-learning algorithms and supervised learning traverse different trajectories in the space of probabilistic models during training but learn similar models eventually; the trajectory of episodic meta-learning for a small ``way'' is about 40$\times$ longer in terms of its Riemann length than that of supervised learning.
\item Contrastive and semi-supervised learning methods traverse similar trajectories to that of supervised learning in the space of probabilistic models.
\item Fine-tuning a model upon a sub-task does not change the representation much if the model was trained
for a large number of epochs.
\end{enumerate}

We present evidence and analysis of these findings using multiple neural architectures and a large number of different image-classification tasks created from the CIFAR-10 and Imagenet datasets.


\section{Methods}
\label{s:methods}

\paragraph{Modeling the task}
We define a task $P$ as a joint distribution on inputs $x \in \reals^d$ and outputs $y \in \cbr{1, \ldots, C}$ corresponding to $C$ classes. Suppose we have $N$ independent and identically distributed samples $\cbr{(x_n, y_n^*)}_{n=1}^N$ from $P$. Let $\yvec = (y_1, \ldots, y_N)$ denote any sequence of outputs on these $N$ samples and $\yvecs$ denote the sequence of ground-truth labels. We can model the task as
\beq{
    P_w(\yvec) = \prod_{n=1}^N p_w^n(y_n)
    \label{eq:def:Pw}
}
where $w$ are the parameters of the model and we have used the shorthand $p_w^n(y_n) \equiv p_w(y_n \mid x_n)$. Let "truth" or $\Ps \equiv P(\yvecs)$ denote the true probability distribution which corresponds to the ground-truth labels. Let "ignorance" or $\Pz$ denote the probability distribution that corresponds to $p^n(y) = 1/C$ for all $n$ and all $y \in \cbr{1,\ldots,C}$.

\paragraph{Bhattacharyya distance}
Given two models $P_u$ and $P_v$ parameterized by weights $u$ and $v$ respectively, the Bhattacharyya distance~\citep{bhattacharyya1946measure} between them averaged over samples can be written as (see~\cref{s:bhatt})
\beq{
    \aed{
        \dB(P_u, P_v)
        &:= -N^{-1} \log \sum_\yvec \prod_n \sqrt{p_u(y_n)\ p_v(y_n)}\\
        &= -N^{-1} \sum_n \log \sum_c \sqrt{p_u^n(c)\ p_v^n(c)}.
    }
    \label{eq:dB}
}
Our model~\cref{eq:def:Pw} involves a product over the probabilities of $N$ samples. Many distances, e.g., the Hellinger distance $2\rbr{1-  \prod_n \sum_c \sqrt{p_u^n(c)\ p_v^n(c)}}$, saturate for large $N$, this is because random high-dimensional vectors are nearly orthogonal. This makes it difficult to use such distances to understand high-dimensional probabilistic models. Bhattacharyya distance is well-behaved for large $N$ due to the logarithm~\citep{quinn2019visualizing,teohVisualizingProbabilisticModels2020}, and that is why it is well suited to our problem.

\begin{remark}[Models with different intermediate representations can have zero Bhattacharyya distance]
Two models can have different internal representations and yet define identical probabilistic models. For example, a representation and a rotated version of the same representation can define identical probabilistic models if this rotation is undone before the output. The Bhattacharyya distance~\cref{eq:dB} only depends on the output probabilities and would be zero if the probabilistic models are identical. Focusing the theory on the probabilistic model that makes the predictions as opposed to the feature space therefore allows us to capture many symmetries in the prediction space.
\end{remark}

\paragraph{Distances between trajectories of probabilistic models}
Consider a trajectory $\rbr{w(k)}_{k =0,\ldots,T}$ that records the weights after $T$ updates of the optimization algorithm, e.g., stochastic gradient descent. This trajectory corresponds to a trajectory of probabilistic models $\tt_w = (P_{w(k)})_{k =0,\ldots,T}$. We are interested in calculating distances between such training trajectories. First, consider $\tt_u = (u(0), u(1), u(2), \ldots, u(T))$ and another trajectory $\tt_v \equiv (u(0), u(2), u(4), \ldots, u(T), u(T), \ldots, u(T))$ which trains twice as fast but to the same end point. If we define the distance between these trajectories as, say, $\sum_k \dB(P_{u(k)}, P_{v(k)})$, then the distance between $\tt_u$ and $\tt_v$ will be non-zero---even if they are fundamentally the same. This issue is more pronounced when we calculate distances between training trajectories of different tasks. It arises because we are recording each trajectory using a different time coordinate, namely its own training progress.

To compare two trajectories correctly, we need a notion of time that can allow us to uniquely index any trajectory. The geodesic between the start point $\Pz$ and the true distribution $\Ps$ is a natural candidate for this purpose since it is unique. Geodesics are locally length-minimizing curves in a metric space. For the product manifold in~\cref{eq:def:Pw}, we can obtain a closed-form formula for the geodesic by noticing that for each sample, the vector $\rbr{\sqrt{p^n_u(c)}}_{c=1,\ldots,C}$ lies on a $C$-dimensional unit sphere. The geodesic connecting two models $P_u$ and $P_v$ under the Fisher information metric which is induced by the Bhattacharyya distance is just the great circle on the sphere~\citep[Eq.~47]{ito2020stochastic}:
\beq{
    \scalemath{0.9}{
    \sqrt{P^\l_{u,v}} = \prod_{n=1}^N \rbr{
    \f{ \sin\rbr{(1 - \l)\dG^n}}{\sin \rbr{\dG^n}} \sqrt{p^n_u} + \f{\sin\rbr{\l \dG^n}}{\sin \rbr{\dG^n}} \sqrt{p^n_v}},
    }
    \label{eq:P_lambda}
}
where $\l \in [0,1]$ and $\dG^n = \cos^{-1} \rbr{\sum_c \sqrt{p_u^n(c)} \sqrt{p_v^n(c)}}$ is one half of the great-circle distance between $p^n_u(\cdot)$ and $p^n_v(\cdot)$. Any probabilistic model $P_w$ on a trajectory $\tt_w$ can now be \textbf{re-indexed by a new ``time'' that we call ``progress''}:
\beq{
    [0,1] \ni t_w = \arg \inf_{\l \in [0,1]} \dG(P_w, P^\l_{0,*}).
    \label{eq:tw}
}
It indicates the distance of $P_w$ to the truth $\Ps$ measured in terms of the closest point $P^{t_w}_{0,*}$ on the geodesic to $P_w$. We solve~\cref{eq:tw} using bisection search~\citep{brent1971algorithm}.
Observe that using the same expression as~\cref{eq:P_lambda}, we can also interpolate between two successive recorded points $P_{w(k)}$ and $P_{w(k+1)}$ of a trajectory by calculating $P^\l_{w(k), w(k+1)}$ for different values of $\l \in [0,1]$. This is useful because different networks train with very different speeds on different tasks, especially in early stages of training. This allows us to effectively convert a sequence of models $\tt_w = (P_{w(k)})_{k=0,\ldots,T}$ into a continuous curve $\t_w = (P_{w(t)})_{t \in [0,1]}$. We calculate the distance between continuous curves $\t_u, \t_v$ as
\beq{
    \dtraj(\t_u, \t_v) = \int_0^1 \dB(P_{u(t)}, P_{v(t)}) \dd{t};
    \label{eq:dtraj}
}
which is approximated using a uniform grid on $[0,1]$.

\paragraph{Riemann length of a trajectory}
Divergences like the Bhattacharyya distance or the Kullback-Leibler (KL) divergence (which is the cross-entropy loss up to a constant)
can be used to define a Riemannian structure in the space of probabilistic models~\citep{amariinformationgeometryits2016}. The distance between two infinitesimally different models $P_w$ and $P_{w+\dd{w}}$ is
\[
    \dd{s}^2 = 4 \dB(P_w, P_{w+\dd{w}}) = \agr{\dd{w}, g(w) \dd{w}} + \oo(\norm{\dd{w}}^2),
\]
where $g(w) = N^{-1} \sum_{\yvec} (P_w)^{-1} \partial^2 P_w$ is the Fisher Information Matrix~\citep[Section A.3]{quinn2019patterns}. This Fisher Information Matrix (FIM) is therefore the metric of the space of the probability distributions and weights $w$ play the role of the coordinates in this space. Up to a scalar factor, the Bhattacharyya distance and the KL-divergence induce the same FIM. The Riemann length of a trajectory $\t_w$ is the integral of these infinitesimal lengths:
\beq{
    \text{Length}(\t_w) = 2 \int_0^1 \sqrt{\dB(P_{w(t)}, P_{w(t+\dd{t})})};
    \label{eq:length}
}
it is equal to the integral of FIM-weighted incremental distance traveled in the weight space. Observe that we do not need the FIM to calculate the length. We can think of the length of a trajectory taken by a model to reach the solution $P_*$ compared to the length of the geodesic as a measure of the inefficiency of the training procedure since the geodesic is the curve with the shortest length. This inefficiency can arise because: (a) not all probability distributions along the geodesic can be parametrized by our model class (approximation error), and (b) the training process may take steps that are misaligned with the geodesic (e.g., due to the loss function, mini-batch updates, supervised vs.\@ some other form of representation learning, etc.).

\paragraph{Mapping a model trained on one task to another task using ``imprinting''}
In this paper, we will consider different tasks $\{P^k \}_{k=1,\ldots,}$ with the same input domain but possibly different number of classes $C^k$. Given a model $P_w^1$ parametrized by weights $w$ for task $P^1$, we are interested in evaluating its learned representation on another task, say, $P^2$. Let $w = (\wb, \wc)$ be the weights for the backbone and the classifier respectively. The logits are $\reals^{C^1} \ni \wc^\top \varphi(x;\wb)$ corresponding to an input $x$ and features of the penultimate layer $\varphi(x;\wb)$. The network's output $p_w(c \mid x_n)$ for $c=1,\ldots,C^1$ is computed using a softmax applied to the logits. If we have learned $w$ from one task $P^1$, then we can re-initialize each row of the classifier weights $(\wc)'_c$ for $c = 1,\ldots, C^2$ to maximize the cosine similarity with the average feature of samples from task $P^2$ with ground-truth class $c$:
\beq{
    \textstyle (\wc)'_c = h/\norm{h}_2 \quad \text{where}\ h = \sum_{\cbr{x: y^*_x = c}} \varphi(x; \wb).
    \label{eq:imprinting}
}
The new network $w = (\wb, \wc')$ can be used to make predictions on $P^2$.
Using imprinting, we can map a trajectory $\t_w^1$ of a network being trained on $P^1$ to another task $P^2$ by mapping each point along the trajectory; let us denote this mapped trajectory by $\tot_w$.

\begin{remark}[Imprinting versus training the final layer or probing]
There are many ways of performing such a mapping, e.g., one could fine-tune the weights using data from $P^2$, linear probing~\citep{shi2016does}, etc. The technique described above is known as ``imprinting''~\citep{huDeepTransferMetric2015,qiLowshotLearningImprinted2018,dhillon2019a}. In this paper, we will be mapping thousands of models across different trajectories to other tasks. Training the final layer, or a new classifier, for all these models is cumbersome and imprinting provides a simple way around this issue.  Note that imprinting is not equivalent to training the classifier $\wc$ (with backbone $\wb$ fixed) using samples from the other task but we found that imprinted weights work well in practice (see \cref{s:app:imprinting}).

%
\end{remark}

\paragraph{How to choose an appropriate task to map different models to?}
Consider the training trajectory $\t^1_u$ of a model being trained on $P^1$ and another trajectory $\t^2_v$ of a model being trained on $P^2$. Using~\cref{eq:imprinting}, we can map these trajectories to the other task to get $\tot_u$ and $\tto_v$. This allows us to calculate, for instance, $\dtraj(\tot_u, \t^2_v)$ using~\cref{eq:dtraj} which is the distance of the trajectory of the model trained on $P^1$ and then mapped to $P^2$ with respect to the trajectory of a model trained on task $P^2$. If the two learning tasks $P^1$ and $P^2$ are very different,  (e.g., Animals in CIFAR-10 and Vehicles in CIFAR-10), then this distance will be large.

Quantities like $\dtraj(\tot_u, \t^2_v)$ or $\dtraj(\tto_v, \t^1_u)$ are reasonable candidates to study similarities between tasks $P^1$ and $P^2$, but they are not equal to one another. We are also interested in doing such calculations with models trained on many different tasks, and mapping them to each other will lead to an explosion of quantities. To circumvent this, we map to a unique task whose output space is the union of the output spaces of the individual tasks, e.g., to study $P^1$ (Animals) and $P^2$ (Vehicles), we will map both trajectories to $\PU$ which is all of CIFAR-10. We will use
\beq{
    \dtraj(\tou_u, \ttu_v)
    \label{eq:dtraj_tou_ttu}
}
as the distance between trajectories trained on $P^1$ and $P^2$.

\paragraph{Visualizing a high-dimensional probabilistic model in lower-dimensions}
We use a visualization technique called intensive principal component analysis (InPCA)~\citep{quinn2019visualizing} that embeds a probabilistic model into a lower-dimensional space. For $m$ probability distributions, consider a matrix $D \in \reals^{m \times m}$ with entries $D_{uv} =  \dB(P_u, P_v)$ and
\beq{
        W = -L D L/2
        \label{eq:w}
}
where $L_{ij} = \delta_{ij} - 1/m$ is the centering matrix. An eigen-decomposition of $W = U \S U^\top$ where the eigenvalues are sorted in descending order of their magnitudes $\abs{\S_{00}} \geq \abs{\S_{11}} \geq \ldots$ allows us to compute the embedding of these $m$ probability distributions into an $m$-dimensional space as $\reals^{m \times m} \ni X = U \sqrt{\S}$. Unlike standard PCA where eigenvalues are non-negative, eigenvalues of InPCA can be both positive and negative, i.e., the lower-dimensional space is a Minkowski space~\citep{quinn2019visualizing}. This allows the InPCA embedding to be an isometry, i.e., pairwise distances are preserved:
\beq{
        \textstyle \sum_{i=1}^m (X_u^i - X_v^i)^2 = \dB(P_u, P_v) \geq 0
        \label{eq:minkowski_distance}
}%
for embeddings $X_u, X_v$ of two distributions $P_u, P_v$. We can measure how well pairwise distances are preserved by a $k$-dimensional sub-space using the ``explained stress'' $\chi_k$~\citep{cox2008multidimensional}:
\beq{
    \textstyle \chi_k
    = 1 - \frac{\norm{W - \sum_{i=1}^k \S_{ii}\ U_i U_i^\top}_{\text{F}}}{\norm{W}_{\text{F}}}
    = 1 - \sqrt{\frac{\sum_{i=k+1}^m \S_{ii}^2}{\sum_{i=1}^m \S_{ii}^2}}.
    \label{eq:explained_stress}
}
Just like standard PCA, if we preserve all the eigenvectors (i.e., $k=m$), then~\cref{eq:minkowski_distance} holds exactly and $\chi_k = 1$. But if we use fewer eigenvectors then pairwise distances can be distorted. See~\cref{s:explained_stress} for more details of the explained stress. \cref{app:inpca_extremely_high_dimensions} describes how we implement InPCA for high-dimensional probabilistic models.


\section{Results}
\label{s:results}

We next describe our findings using the theoretical ideas developed in the previous section. We present a broad range of evidence and analysis using a large number of representation learning techniques, multiple neural architectures and a large number of different image-classification tasks created from the CIFAR-10 and ImageNet datasets. Experiments in this paper required about 30,000 GPU-hours. \cref{s:app:setup} describes the setup for these experiments in detail. See~\cref{s:faq} for a discussion of some frequently asked questions. One more result, Result 7: Contrastive learning methods trained on different datasets learn similar representations, is presented in~\cref{s:app:results}.

\begin{remark}
All the analysis in this paper (except~\cref{fig:meta_learning_panel1,fig:meta_learning_panel2}) was conducted using the test data. All models were trained using the training data, but all mapped models, distances between trajectories, quantitative evaluation of progress and InPCA embeddings were computed using the test dataset. The reason for this is that we would like to study the geometry of tasks as evidenced by samples that were not a part of training. To emphasize, we do not develop any new algorithms for learning in this paper. Therefore using the test data to quantify relationships between tasks is reasonable; see similar motivations in~\citet{kaplun2022deconstructing} or ~\citet{ilyas2022datamodels} among others. Our findings remain valid when training data is used for analysis; this is because in most of our experiments, a representation is trained on one task but makes predictions on a completely new task after mapping.
\end{remark}

\begin{figure}[!htb]
\centering
\includegraphics[width=0.8\linewidth]{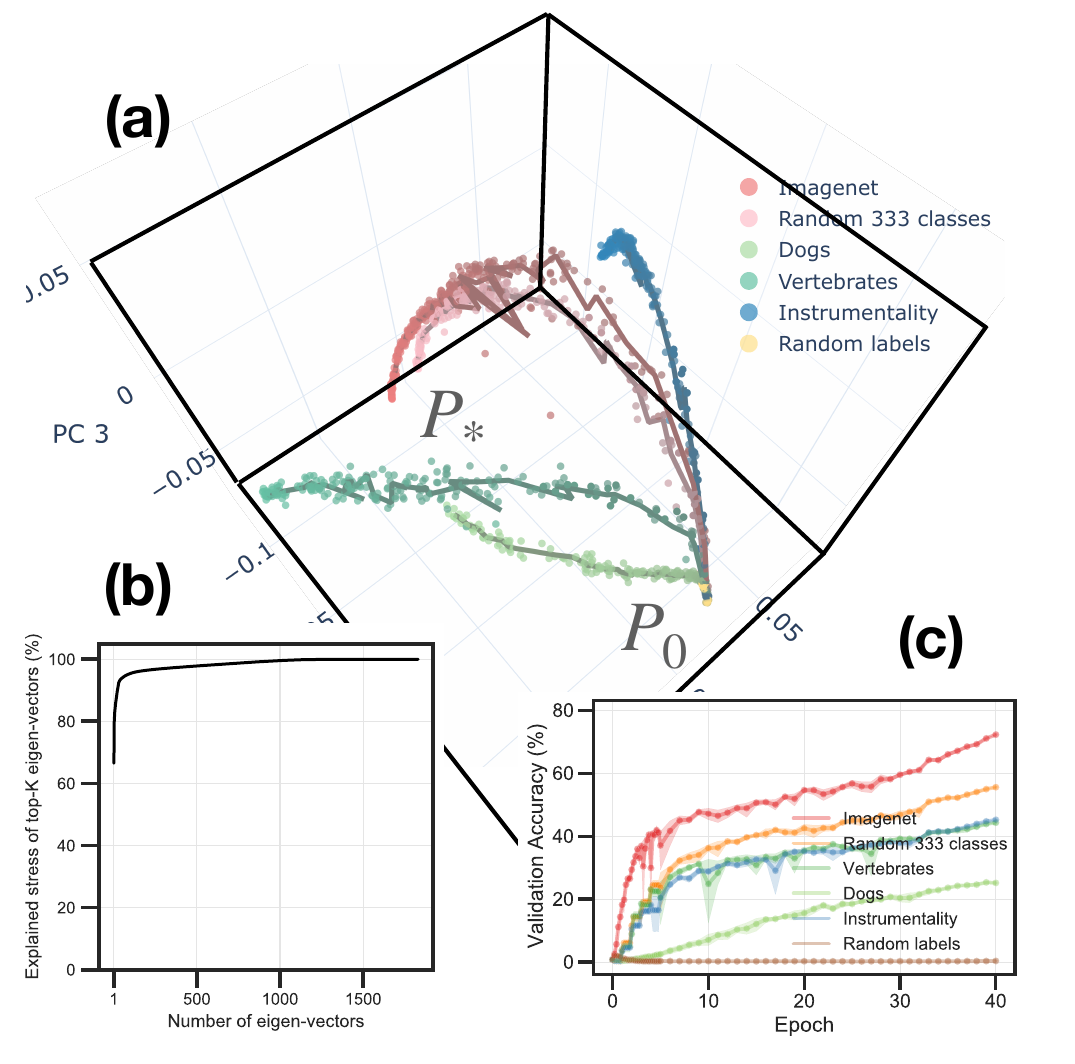}
\caption{
\textbf{(a)} Visualization of training trajectories of models trained on 6 tasks from ImageNet. Each point is one network, bold lines connect points along the average trajectory of each task (across 5 random weight initializations). Trajectories move towards the truth $P_*$, which corresponds to the ground-truth labels. Training on one task makes a remarkable amount of progress on unseen, seemingly dissimilar, classes. Trajectories of models trained on a random set of 333 classes are similar to those of the entire ImageNet. Some classes (Instrumentality) are closer to this trajectory while others such as Vertebrates and Dogs are farther away. Dogs is a semantic subset of Vertebrates; it splits at the beginning but seems to eventually reach a similar representation as one of the intermediate points of Vertebrates.\\[0.25em]
\textbf{(b)} Percentage explained stress~\cref{eq:explained_stress} captured by subspace spanned by the top $k$ InPCA eigenvectors.\\[0.15em]
\textbf{(c)} Validation accuracy on different tasks vs.\@ epochs.\\[0.5em]
}
\label{fig:imagenet1}
\end{figure}

\begin{figure}[!htb]
\centering
\includegraphics[width=\linewidth]{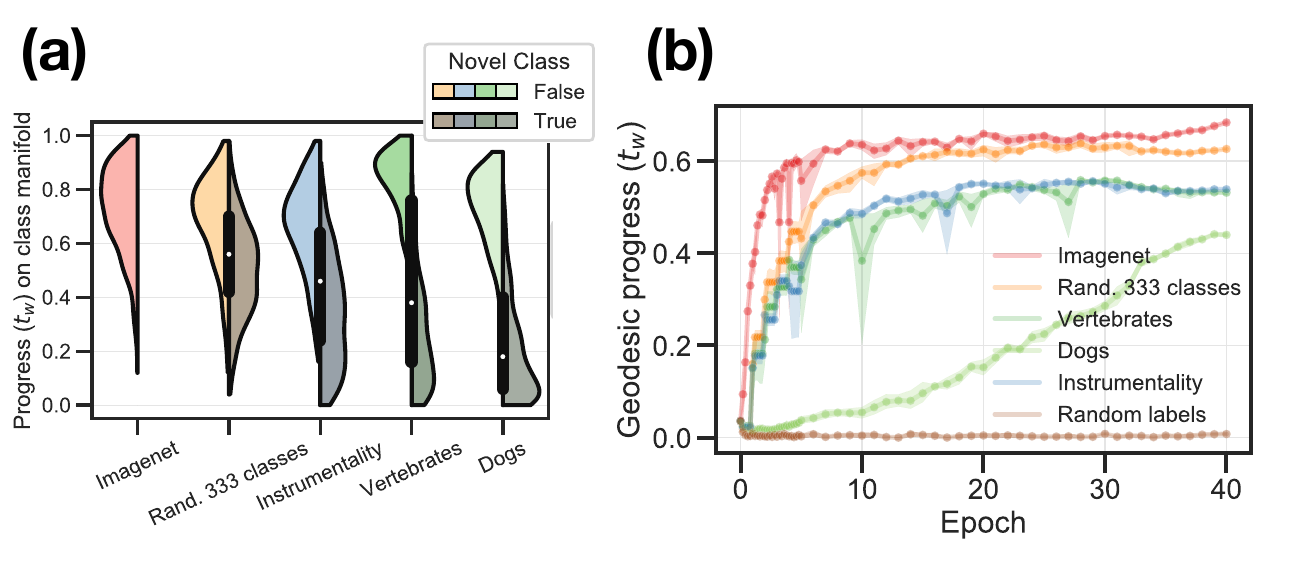}
\caption{
\textbf{(a)} Progress made by each model on classes seen during training (left half, lighter shade) and on novel classes (right half, darker shade). We compute $t_w^c$ which is the progress $t_w$ of images restricted to a single class $c$. This quantity $t_w^c$ measures the quality of the representation for class $c$. Violin plots denote the distribution of $t_w^c$ indicate that we make more progress on classes seen during training. If the model sees a larger diversity of classes (like with random 333 classes), more progress is made on the novel classes. Surprisingly, even if we train on just the ``Dogs", we make some progress on novel classes.\\[0.25em]
\textbf{(b)} Progress $t_w$~\cref{eq:tw} on the Y-axis against the number of epochs of training on the X-axis. The progress $t_w$ increases with more epochs of training---all models make non-trivial progress towards the truth $P_*$ ($t_w =1$). Even if we train on only Dogs (118 classes) we make progress on the entire ImageNet.\\[0.1em]
}
\label{fig:imagenet2}
\end{figure}

\paragraph{Result 1: The manifold of models trained on different tasks, and using different representation learning methods, is effectively low-dimensional}
We trained multiple models on 6 different sub-tasks of ImageNet (from 5 random initializations each) to study the dimensionality of the manifold of probabilistic models along the training trajectories (100 points equidistant in progress~\cref{eq:tw}) after mapping all models to all ImageNet classes ($\sim 10^8$ dimensions). We use the explained stress (defined in~\cref{s:explained_stress}), to measure if the distances are preserved by the first $k$ dimensions of the embedding of the models. The first 3 dimensions of InPCA (\cref{fig:imagenet1}a) preserve 80.02\% of the explained stress (\cref{fig:imagenet1}b shows more dimensions). This is therefore a remarkably low-dimensional manifold. It is not exactly low-dimensional because the explained stress is not 100\%, but it is an effectively low-dimensional manifold. This also indicates that the individual manifolds of models trained on one task are low-dimensional, even if they start from different random initializations in the weight space. Such low-dimensional manifolds are seen in \emph{all} our experiments, irrespective of the specific method used for representation learning, namely, supervised, transfer (fine-tuning), meta, semi-supervised and contrastive learning.

\begin{remark}[A detailed description of how we plot trajectories of representations]
We provide a non-mathematical description of how the theory in~\cref{s:methods} was used to draw~\cref{fig:imagenet1}a below. We train 5 different networks (random seeds for initialization) for each of the 6 tasks, and record 61 model checkpoints during training; this gives 1830 checkpoints for this experiment. We re-index all checkpoints to calculate their progress using~\cref{eq:P_lambda,eq:tw}. We then interpolate between each consecutive pair of the 61 checkpoints along each trajectory using~\cref{eq:P_lambda}. The training trajectory can now be sampled at any progress $t_w \in [0,1]$. We next calculate the ``average trajectory'' of the 5 networks (random seeds) of each task by averaging the output probabilities in~\cref{eq:def:Pw} at a fixed value of $t_w$; 100 different values of $t_w$ spread uniformly between $[0,1]$ are chosen. These 100 points along the average trajectory of each of the 6 tasks are also embedded together with the 1830 checkpoints (i.e., $m=$ 2430 in~\cref{eq:w}). \cref{fig:imagenet1}a plots the top three dimensions obtained from InPCA. To clarify, the explained stress of the top 2430 dimensions would be exactly 100\%.
\end{remark}

\begin{figure}[!t]
\centering
\includegraphics[width=\linewidth]{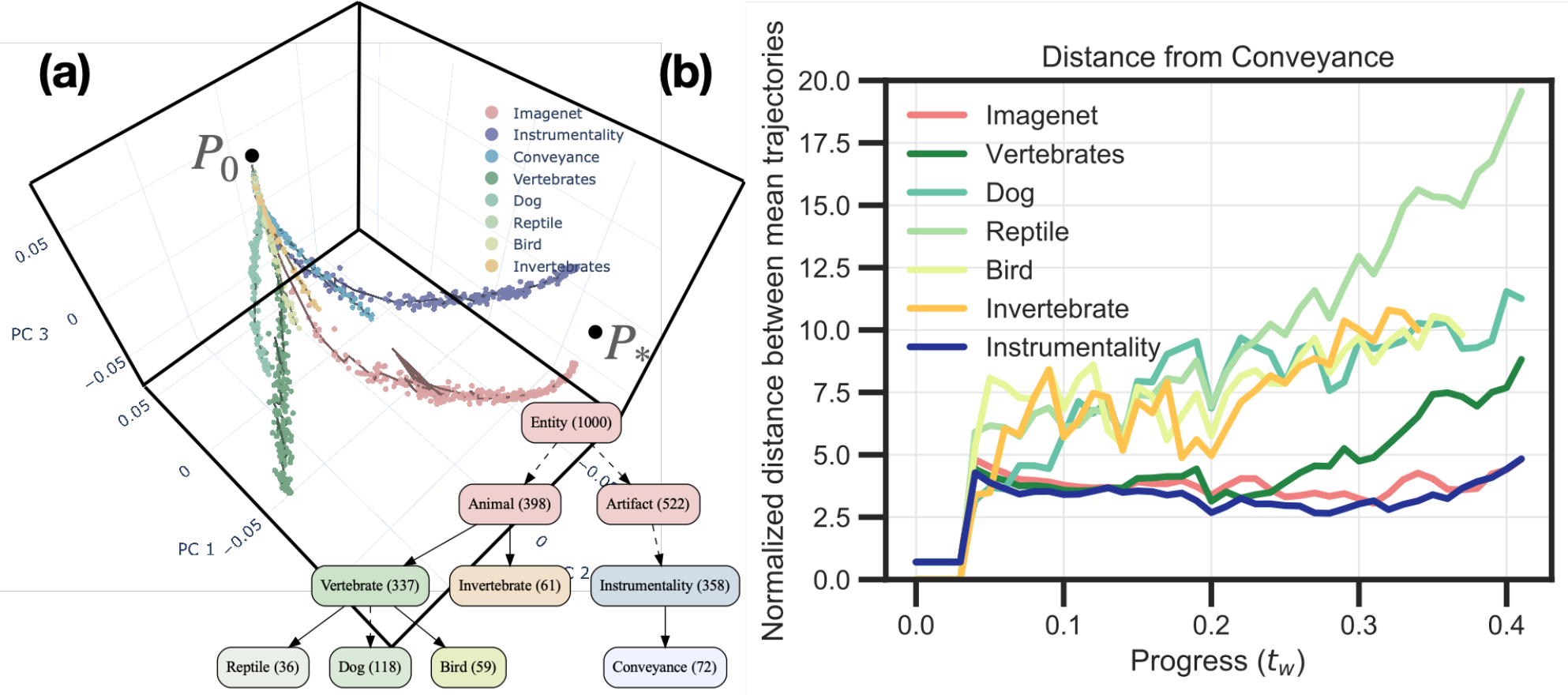}
\caption{
\textbf{(a)} Trajectories of models trained on different phyla of Wordnet (inset). The model manifold is again effectively low-dimensional (78.72\% explained stress in 3 dimensions).\\[0.2em]
\textbf{(b)} We analyze the trajectories in~\cref{fig:result3}(a) and obtain a quantitative description of how trajectories of different tasks diverge from each other during training; the procedure is explained in in~\cref{rem:interp_radii_plots}. The plot depicts the Bhattacharyya distance between the mean trajectories (over random initializations) on different tasks, and the mean trajectory of Conveyance. This distance is normalized by the average of the tube radii (maximum distance of one of the 5 trajectories from the mean, computed at each progress) of the two trajectories. Such quantities allow us to make precise statements about the differences between representations and show some very surprising conclusions. Trajectories of tasks that are nearby in Wordnet are also nearby in terms of their learned representations. Further, trajectories of ImageNet (pink) are closer to Conveyance (as expected), but those of Vertebrates (red) are equally far away for more than 60\% ($t_w \approx 0.25$) of the progress. In other words, training on Vertebrates (reptiles, dog, bird) makes a remarkable progress on Conveyance (cars, planes).\\
}
\label{fig:result3}
\end{figure}

\begin{figure}[!ht]
\centering
\includegraphics[width=0.99\linewidth]{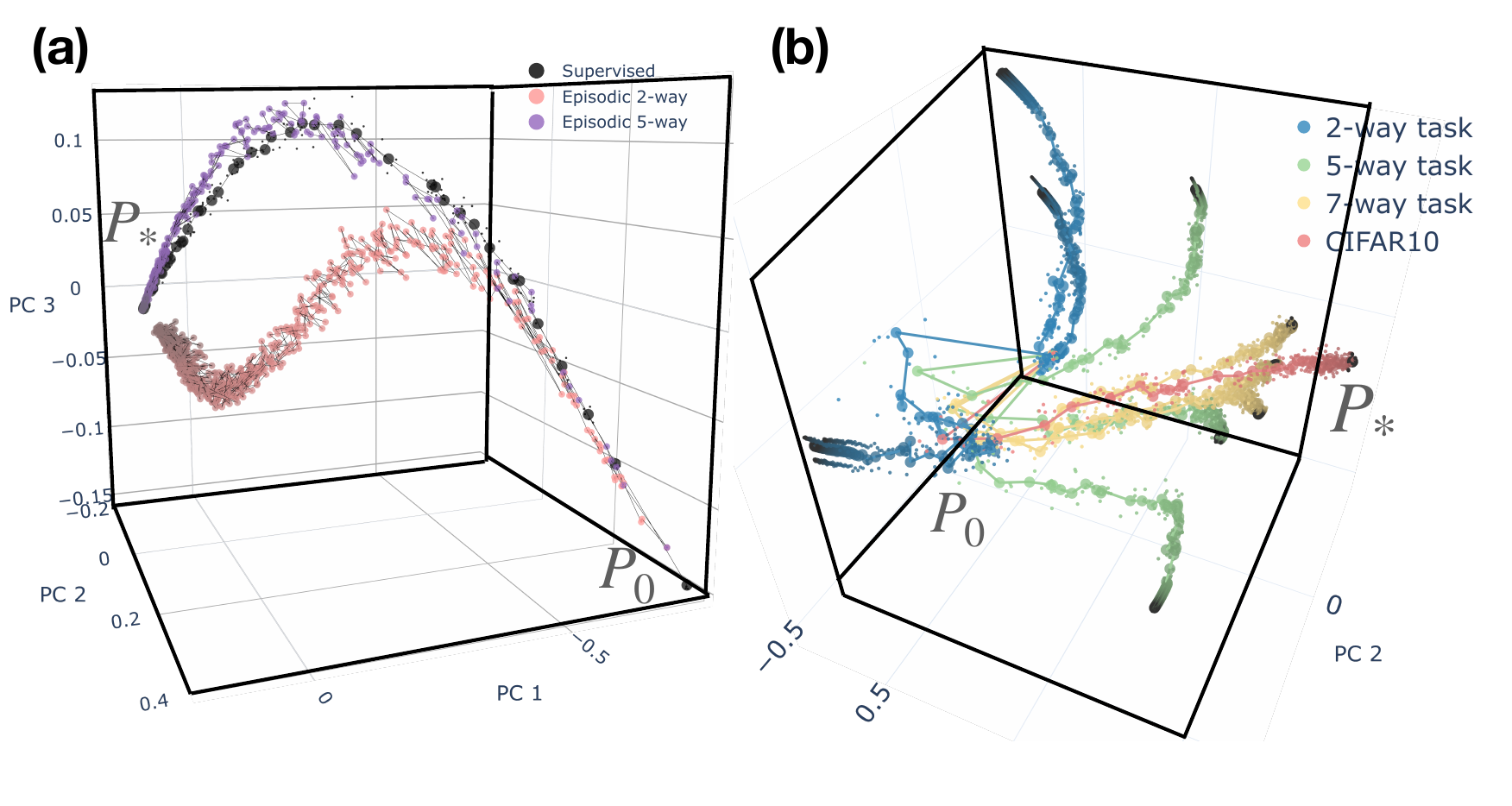}
\caption{
\textbf{(a)} Training trajectories for supervised learning (black), 2-way (pink) and 5-way episodic meta-learning (purple). Trajectories of 5-way meta-learning are very similar to those of supervised learning and eventually reach very similar models and high test accuracy. In contrast, 2-way meta-learning has a much longer trajectory (about 40$\times$ longer in Riemann length than black) and does not reach a good test accuracy (on all 10 CIFAR-10 classes). Representations are similar during early parts of training even if these are quite different learning mechanisms.\\[0.15em]
\textbf{(b)} Trajectories of 2-way (blue), 5-way (green), 7-way (yellow) tasks trained using cross-entropy loss compared to supervised learning (red). For large ``way'', trajectories are similar to supervised learning but they quickly deviate from the red trajectories for small ways.\\[0.15em]
}
\label{fig:meta_learning_panel1}
\end{figure}

\begin{figure}[!thb]
\centering
\includegraphics[width=0.9\linewidth]{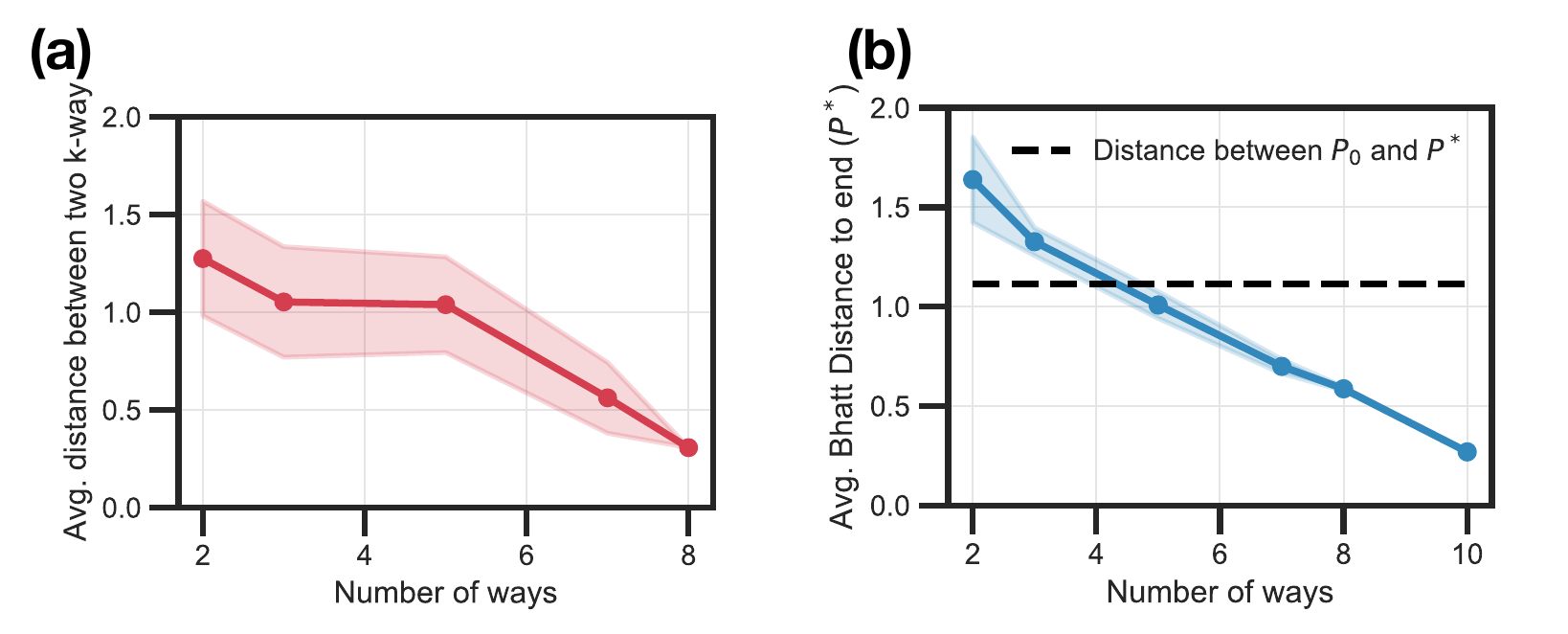}
\caption{
\textbf{(a)} Average distance between two $k$-way meta-learning trajectories decreases with $k$, this is a geometric evidence of the variance of predictions of learned representations.\\[0.15em]
\textbf{(b)} Training with a small way leads to models that predict poorly on test data (large distances from truth). These embeddings were calculated using the training dataset. The rationale being that we wanted to show how different meta-learning and supervised learning are during training. \\
}
\label{fig:meta_learning_panel2}
\end{figure}

\paragraph{Result 2: Supervised learning on one task results in a surprising amount of progress on seemingly dissimilar tasks. Progress on other tasks is larger if the training task has diverse classes.}
We studied the progress $t_w$~\cref{eq:tw} made by models (\cref{fig:imagenet2}b) trained on tasks from Result 1. Training on the task ``Dogs'' makes non-trivial progress on other tasks, even seemingly dissimilar ones like ``Instruments'' which contains vehicles, devices and clothing. In fact, it makes a remarkable amount of progress on the \emph{entire} ImageNet, about 63.38\% of the progress of a model trained directly on ImageNet. Progress is larger for larger phyla of ImageNet (Vertebrates and Instruments). But what is surprising is that if we train on a random subset of 333 classes (a third of ImageNet), then the progress on the entire ImageNet is very large (92\%). This points to a strong shared structure among classes even for large datasets such as ImageNet. Note that this \emph{does not} mean that tasks such as Vertebrates and Instruments are similar to each other. Even if training trajectories are similar for a while, they do bifurcate eventually and the final models are indeed different (see~\cref{fig:result3}b and~\cref{rem:interp_radii_plots} on how to interpret it).

In~\cref{fig:imagenet2}a, we studied the projections of models trained on one task onto the geodesics of unseen classes calculated using~\cref{eq:P_lambda} evaluated at the progress $t_w$~\cref{eq:tw}). We find that a model trained on the entire ImageNet makes uneven progress on the various classes (but about 80\% progress across them, progress is highly correlated with test error of different classes). Models trained on the 6 individual tasks also make progress on other unseen classes. As before, training on Instruments, Vertebrates, Dogs makes smaller progress on unseen classes compared to training on a random subset of 333 classes. This is geometric evidence that the more diverse the training dataset, the better the generalization to \emph{unseen} classes/tasks; this phenomenon has been widely noticed and utilized to train models on multiple tasks, as we discuss further in Result 4.

\paragraph{Result 3: The structure of the space of tasks indicated by our visualization technique is consistent with parts of the Wordnet phylogenetic tree.}
To obtain a more fine-grained characterization of how the geometry in the space of learnable tasks reflects the semantics of these tasks, we selected two particular phyla of ImageNet (Animals, Artifacts) and created sub-tasks using classes that belong to these phyla (\cref{fig:result3}a). Trajectories of models trained on Instruments and Conveyance are closer together than those of Animals. Within the Animals phylum, trajectories of Vertebrates (Dog, Reptile, Bird) are closer together than those of Invertebrates (\cref{fig:result3}b for quantitative metrics). Effectively, we can recover a part of the phylogenetic tree of Wordnet using our training trajectories. We speculate that this may point to some shared structure between visual features of images and natural language-based semantics of the corresponding categories which was used to create Wordnet~\citep{miller1998wordnet} of the corresponding categories. Such alignment with a natural notion of relatedness also demonstrates the soundness and effectiveness of our technical machinery.

\begin{remark}[Building a precise and quantitative characterization of trajectories of representations]
\label{rem:interp_radii_plots}
The precise way to understand statements like those in Result 3 is using the quantitative analysis reported in~\cref{fig:result3}b and~\cref{fig:app:trajectory_distances}. To expand upon the caption, the X-axis of the plot is progress. For multiple models (5 random seeds) trained on two tasks (say Conveyance and Dogs), we have calculated the mean (across random seeds) of the interpolated trajectories at different progress. At each specific progress, we have plotted the distance between the mean model trained on Conveyance (say task 1) and Dogs (say task 2) divided by the average tube radii (which is the maximum of the distance of the model corresponding to one seed from the mean):
\[
    \textstyle 2 \dB(\t_{\text{mean}}^{1 \to U}, \t_{\text{mean}}^{2 \to U})/\sum_{k=1,2} \max_a[\dB(\t_a^{k \to U}, \t_{\text{mean}}^{k \to U})].
\]
The is a measure of how far away the trajectories of these two models are. If it is less than 1, then the ``tubes'' corresponding to models trained on tasks 1 and task 2 intersect.

Let us emphasize that we have performed such analyses for all experiments in this paper (see~\cref{fig:app:trajectory_distances}); while the InPCA embedding gives an easy-to-understand visual description of these results for high-dimensional probabilistic models, the information geometric techniques developed in this paper enable us to make these descriptions precise and quantitative. We also include a similar step-by-step guide on how to interpret~\cref{fig:ssl_panel}b in~\cref{s:app:results}.
\end{remark}

\paragraph{Result 4: Episodic meta-learning algorithms traverse very different trajectories during training but they fit a similar model eventually.}
Meta-learning methods build a representation which can be adapted to a new task~\citep{thrun2012learning}. We studied a common variant, the so-called episodic training methods~\citep{bengioOptimizationSynapticLearning1992}, in the context of few-shot learning methods~\citep{vinyalsMatchingNetworksOne2016}. In these methods, each mini-batch consists of samples from $C^w$ out of $C$ classes (called ``way'') split into two parts: a ``support set'' $D_s$ of $s$ samples/class (called ``shot''), and a ``query set'' $D_q$ of $q$ samples/class. Typical methods, say prototypical networks of~\citet{snellprototypicalnetworksfewshot2017}, implement a clustering loss on features of the query samples using averaged features of the support samples $\varphi_c = s^{-1} \sum_{\cbr{x \in D_s, y^*(x) = c}} \varphi(x; \wb)$ for all $c=1,\ldots, C^w$ as the cluster centroids.
If features $\varphi$ lie on an $\ell_2$ ball of radius 1, then doing so is akin to maximizing the cosine similarity between cluster centroids and features of query samples. The same clustering loss with the learned backbone $\wb$ is used to predict on unseen classes (using ``few'' support samples to compute centroids) at test time.

To understand the representations learned by episodic meta-learning methods, we compared trajectories of episodic meta-learning to the trajectory taken by supervised learning in~\cref{fig:meta_learning_panel1}. Supervised learning uses the cross-entropy loss over all the $C$ classes while episodic meta-learning optimizes a loss that considers all k-way classification tasks (where $k$ is typically smaller than $C$), its objective differs from that used for supervised learning. Since the two objectives are different, it comes as a surprise that both arrive at the same solution; see~\cref{fig:meta_learning_panel1}a,b and~\cref{fig:app:meta_dist} for distances between trajectories. But the Riemann trajectory length of episodic training is about 40$\times$ longer than that of supervised learning. It is worth noting that the explained stress is only 40.96\% in~\cref{fig:meta_learning_panel1}a because of larger fluctuations for episodic learning in other directions. Therefore, episodic meta-learning has a qualitatively different training trajectory in the prediction space than supervised learning. The implications of this are consistent with recent literature which has noticed that the performance of few-shot learning methods using supervised learning (followed by fine-tuning) is comparable to, or better than, episodic meta-learning~\citep{dhillon2019a,kolesnikovbigtransferbit2020,fakoor2019meta}. Indeed, a supervised learned representation also minimizes the clustering loss.

In order to understand why few-shot accuracy of episodic training is better with a large way~\citep{gidarisDynamicFewshotVisual2018}, we trained models on different 2-way 5-way and 7-way tasks using the cross-entropy loss (\cref{fig:meta_learning_panel1}b). We find that the radius of the tube that encapsulates the models of 2-way tasks around their mean trajectory is very large, almost as large as the total length of the trajectory, i.e., different models trained with a small way tasks traverse very different trajectories. Tube radius decreases as the way increases (\cref{fig:meta_learning_panel2}a). Further, the distance of models from the truth $P_*$ (which is close to the end point of the supervised learning model) is higher for a small way (\cref{fig:meta_learning_panel2}b). This is geometric evidence of the widely used empirical practice of using a large way in episodic meta-learning. Observe in~\cref{fig:meta_learning_panel2}b that as the way increases, the trajectory becomes more and more similar to that of supervised learning. See~\cref{fig:app:trajectory_distances} for a quantitative analysis of these trajectories.

\begin{figure}[!htb]
\centering
\includegraphics[width=0.48\linewidth]{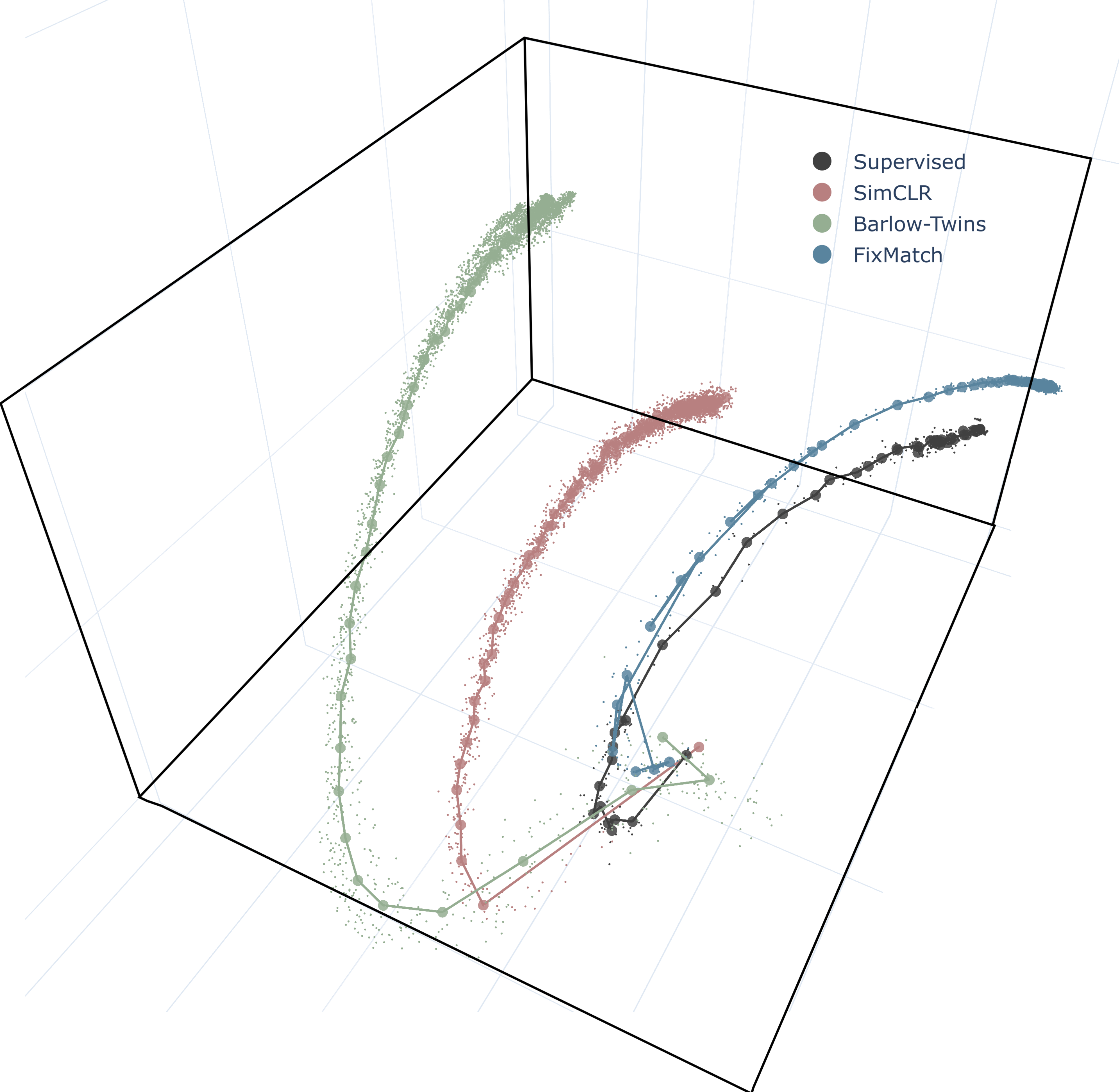}
\includegraphics[width=0.48\linewidth]{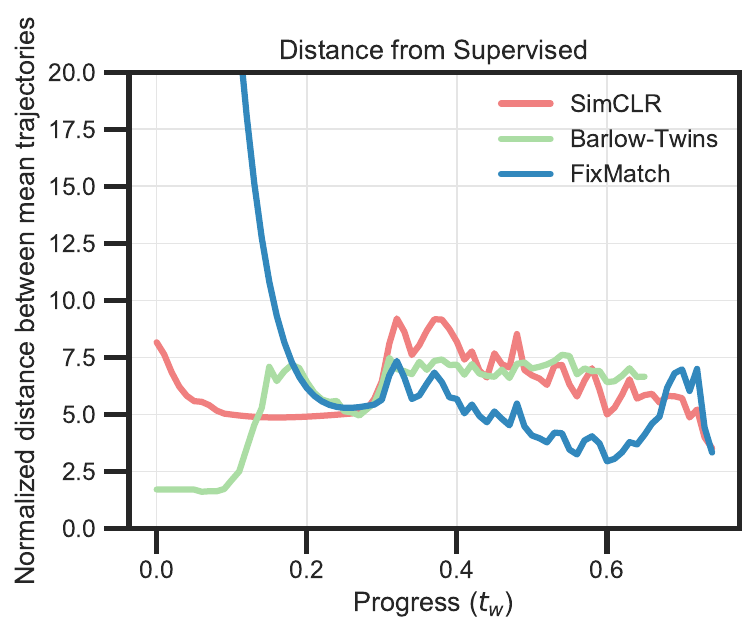}
\caption{
We consider 4 methods for training on CIFAR10: supervised learning, SimCLR~\citep{chen2020simple}, Barlow-twins~\citep{zbontar2021barlow} and Fixmatch~\citep{sohn2020fixmatch}. Fixmatch has access to 2500 labeled samples and 47500 unlabeled samples. SimCLR and Barlow-twins use 50,000 unlabeled samples for training. \\
\textbf{(a)} We plot the trajectories for supervised, semi-supervised and contrastive learning. The trajectory of semi-supervised learning (Fixmatch) eventually resembles supervised learning in comparison to contrastive learning methods. All methods result in remarkably similar trajectories although some of these methods are trained using only unlabeled data.\\
\textbf{(b)} Normalized distance of trajectories corresponding to contrastive and semi-supervised learning to the trajectory of supervised learning. Semi-supervised learning (Fixmatch) deviates considerably from the other methods at the beginning. We speculate that this is because the trajectory of Fixmatch is influenced by the 2500 labeled samples. As, training progresses, Fixmatch becomes increasingly similar to supervised learning as evidenced by the dip in the blue line for larger values of progress $(t_w)$.\\}
\label{fig:semi_ssl}
\end{figure}

\paragraph{Result 5: Contrastive and semi-supervised learning methods traverse trajectories similar to those of supervised learning.}
Contrastive learning~\citep{becker1992self} learns representations without using ground-truth labels~\citep{gutmann2010noise,chen2020simple}. It has been extremely effective for self-supervised learning~\citep{doersch2017multi,kolesnikov2019revisiting}, e.g., prediction accuracy with 1--10\% labeled data is close to that of supervised learning using all data~\citep{chen2020big}. Semi-supervised methods~\citep{berthelot2019mixmatch,sohn2020fixmatch} learn representations when ground-truth labels are available for only a small fraction of the data (0.1--1\%). These methods achieve a prediction accuracy within 5\% of the accuracy achieved through supervised learning. We compared representations learned using contrastive and semi-supervised learning with those from supervised learning to understand why these methods are so effective.

Consider a task $P$ and a set of augmentations $G$ (e.g., cropping, resizing, blurring, color/contrast/brightness distortion etc.). Given inputs (say images) $x$ from $P$, contrastive learning forces the representation $\varphi(g(x); \wb)$ and $\varphi(g'(x); \wb)$ (shortened to $\varphi(g(x))$ below) of the same input for two different augmentations $g, g'$ to be similar. And forces it to be different from representations of other augmented inputs $x'$~\citep{zbontar2021barlow,bachman2019learning,dosovitskiy2014discriminative}. Semi-supervised learning methods have access to both labeled inputs $x_l$ and unlabeled inputs $x_u$. More recent methods are usually trained to fit the labeled inputs using the cross-entropy loss while enforcing consistent predictions across all augmentations~\citep{tarvainen2017mean,berthelot2019mixmatch} for any unlabeled input.

We compare the representations of semi-supervised (Fixmatch~\citep{sohn2020fixmatch}), contrastive (SimCLR~\citep{chen2020simple}, Barlow-twins~\citep{zbontar2021barlow}) and supervised learning in~\cref{fig:semi_ssl}. All three trajectories are similar to the trajectory of supervised learning. We find that the trajectory of semi-supervised learning deviates from the supervised learning trajectory initially, but the two are very similar for larger values of progress ($t_w$). This points to a remarkable ability of semi and self-supervised learning methods to learn representations that are similar to those of supervised learning; it is not just that the accuracy of these methods is similar, they also learn similar probabilistic models.

\paragraph{Result 6: Fine-tuning a pre-trained model on a sub-task does not change the representation much.}
To understand how models train on multiple tasks, we selected two binary classification sub-tasks  of CIFAR-10 (Airplane vs.\@ Automobile, and Bird vs.\@ Cat).

\begin{figure}[!thb]
    \centering
    \includegraphics[width=0.9\linewidth]{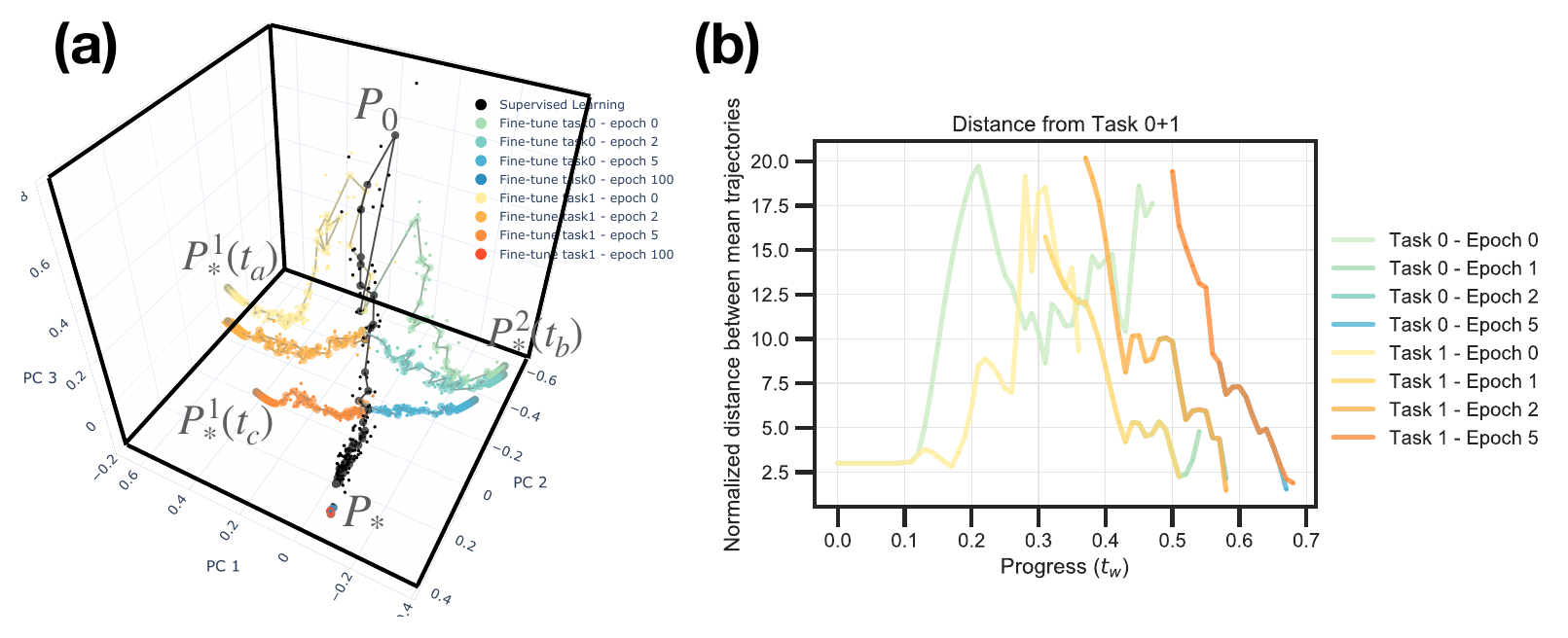}
    \vspace*{-1em}
    \caption{
    \textbf{(a)}
    Fine-tuning trajectories on Airplane vs.\@ Automobile, and Bird vs.\@ Cat sub-tasks of CIFAR-10 (warm and cold hues) pre-trained from different points along the trajectory of supervised learning. If the pretrained model has progressed further towards the truth $P_*$, then fine-tuning it on a sub-task does not change the representation much. The final trajectory (fine-tuning from epoch 100) is indistinguishable from $P_*$.
    \textbf{(b)}
    Bhattacharyya distance between the mean trajectories normalized by the average of the tube radii (like~\cref{fig:result3}b). models (say, fine-tuned after epoch 5 on task 1) go \emph{backwards} in terms of progress, i.e., they unlearn the pre-trained representation in order to fit the new task. This occurs as early as epoch 1 here. It suggests that learning occurs extremely rapidly at the beginning and determines the efficiency of fine-tuning. Some curves here are not visible because they are overlapping heavily. \\}
    \label{fig:fine_tuning_panel}
\end{figure}

\begin{figure}[!htb]
\centering
\includegraphics[width=0.85\linewidth]{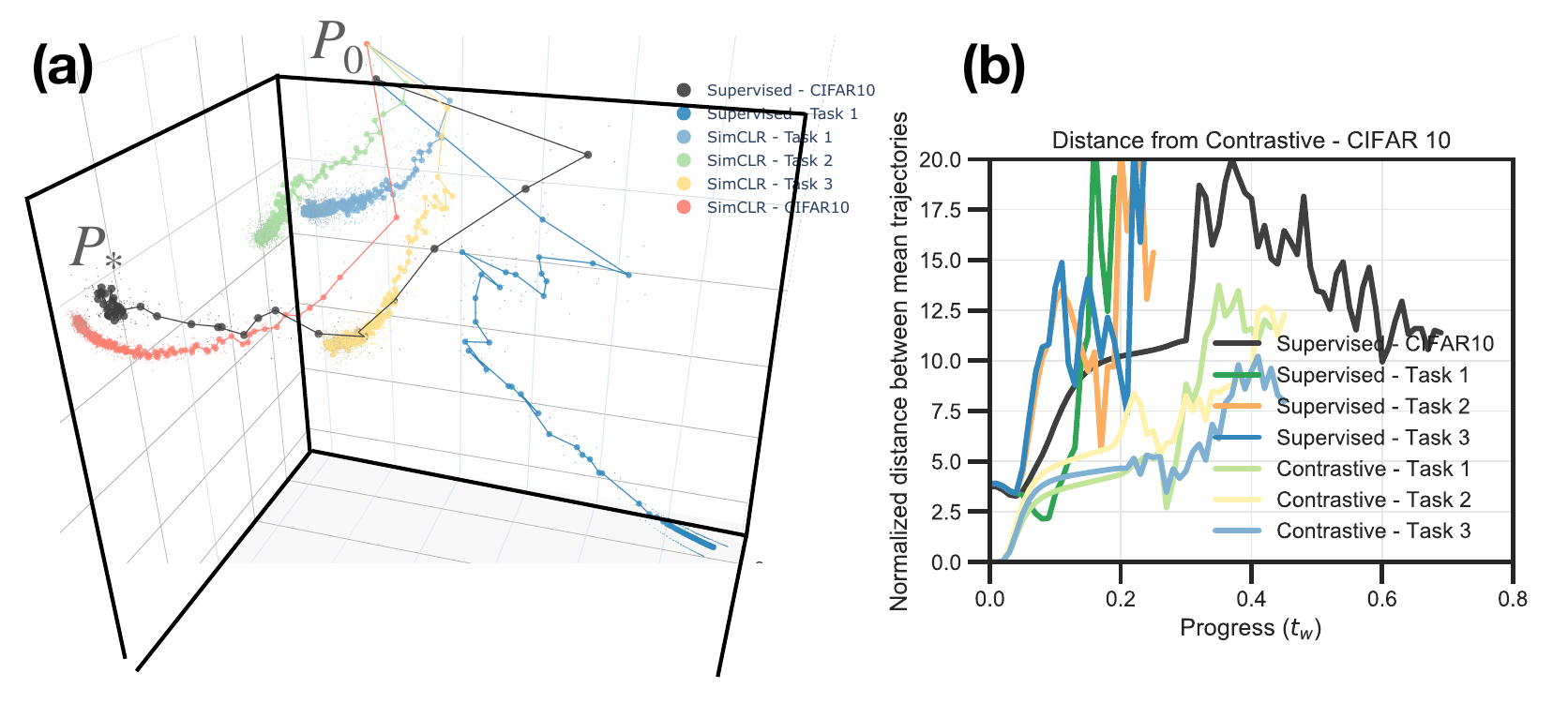}
\caption{
\textbf{(a)} Trajectories of contrastive learning (SimCLR) on 3 datasets (two classes each) and entire CIFAR-10 compared to those of supervised learning. SimCLR on entire CIFAR-10 learns a similar representation as that of the supervised learned model $P_*$ (which fits the training data perfectly). SimCLR trajectories are close to each other even if different datasets were used to train them. It may seem from the embedding that SimCLR trajectories are similar to that supervised learning, which would be very surprising because the former does not use any labels, but see below.\\[0.25em]
\textbf{(b)}
Bhattacharyya distance between the mean trajectories of all models and the mean trajectory of SimCLR on all CIFAR-10. This distance is normalized by the average of the tube radii (like~\cref{fig:fine_tuning_panel}b). SimCLR trajectories of two-class datasets are indeed very close to each other (mean distance is $\sim 5\times$ more than their tube radii for about 45\% of the way ($t_w \approx 0.2$)). This plot indicates that two-class SimCLR trajectory (light blue) is close to SimCLR on all of CIFAR-10. But two-class supervised learning trajectory (darker blue) is much farther away from SimCLR on all of CIFAR-10.
}
\label{fig:ssl_panel}
\end{figure}

We selected models at different stages of standard supervised learning on CIFAR-10 (i.e., using 10-way output and softmax cross-entropy loss) and fine-tuned each of these models on two sub-tasks (the entire network is fine-tuned without freezing the backbone). As~\cref{fig:fine_tuning_panel} shows, models that were fine-tuned from earlier parts of the trajectory travel a large distance and move away from trajectories of the supervised learned CIFAR-10 models. As we fine-tune later and later models, the distance traveled away from the trajectory is smaller and smaller, i.e., changes in the representation are smaller. For a fully-trained CIFAR-10 model which interpolates the training data, the distance traveled by fine-tuning is very small (the points are almost indistinguishable in the picture); this is because both $P^1$ and $P^2$ are subsets of CIFAR-10.

Algorithms for transfer learning train on a source task before fine-tuning the model on the target task. If two tasks share a large part of their training trajectory, then we may start the fine-tuning from many shared intermediate points---there are many such points. If the chosen point is farther along in terms of progress then the efficiency resulting from using the source task is higher because the trajectory required to fit the target task is shorter; such trajectories were used in~\citep{gao2020information} to define a distance between tasks. As we saw in Result 2, trajectories of different tasks bifurcate after a shared part. The resultant deviation less for related tasks and more for dissimilar tasks (\cref{fig:fine_tuning_panel}a,~\cref{fig:imagenet1}a,c). Therefore it is difficult to know \emph{a priori} from which point one should start the fine-tuning from without knowing the manifold of the target task. In particular, our geometric picture indicates that fine-tuning from a fully-trained model can be detrimental to the accuracy on the target task. This has been noticed in a number of places in the transfer learning literature, e.g.,~\citet{li2019rethinking}, and has also been studied theoretically~\citep{gao2020free}. 

\paragraph{Result 7: Contrastive learning methods trained on different datasets learn similar representations}
We compared representations learned using contrastive learning with those from supervised learning to understand some aspects of why the former are so effective.

We used SimCLR~\citep{chen2020simple} to perform contrastive learning on images from four sets of classes (airplane-automobile, bird-cat, ship-truck and all of CIFAR-10). We compared the learned representation to that from supervised learning on two tasks (airplane-automobile and all of CIFAR-10)
in~\cref{fig:ssl_panel}. Models trained using contrastive learning on two-class datasets learn very different representations from models trained on the same task but using supervised learning. Models trained using contrastive learning on different datasets learning similar representations (trajectories of all three two-class datasets are very close to each other). This is reasonable because contrastive learning does not use any information from the labels. It is surprising however that the trajectory of models from contrastive learning on these two-class datasets is similar to trajectories of models from contrastive learning on the entire CIFAR-10.

Let us elaborate upon this a bit more. We have color-matched the lines in~\cref{fig:ssl_panel}b with those in~\cref{fig:ssl_panel}a. The black curve is the trajectory of supervised learning on the entire CIFAR-10; red is the trajectory of SimCLR trained on the entire CIFAR-10. \cref{fig:ssl_panel}b compares the distances of trajectories in~\cref{fig:ssl_panel}a from the red one ``contrastive''; this is why there is no red trajectory in~\cref{fig:ssl_panel}b.

\begin{itemize}[nosep]
\item The first thing to note here is that the black and red trajectories are quite close to each other; the black line in~\cref{fig:ssl_panel}b is only about 20 times far away from red as compared to their corresponding tube radii.
\item Next observe that the trajectory of SimCLR on Task 1 (light blue), SimCLR on Task 2 (green) and SimCLR on Task 3 (yellow) are very similar to each other; this is seen in both~\cref{fig:ssl_panel}a and in~\cref{fig:ssl_panel}b.
\item Third, they are closer to SimCLR on all of CIFAR-10 than any supervised learning trajectories (this is seen in~\cref{fig:ssl_panel}b because their curves are below everyone else). Thus, contrastive learning on datasets with different classes learns similar representations.
\item The learned representation of two-class SimCLR models is similar to the one obtained using data from all classes (red) (in this experiment this occurs up to about $t_w =$ 0.4 progress) but they do not go all the way to the truth (i.e., the end point of black line). This shows the benefit of having data from many classes during contrastive learning.
\end{itemize}
Also see~\cref{fig:app:trajectory_distances} for distances computed with respect to other trajectories which can be used to further investigate these claims.

\section{Related Work and Discussion}
\label{s:related}

\paragraph{Understanding the space of learnable tasks}
A large body of work has sought to characterize relationships between tasks, e.g., domain specific methods~\citep{zamir2018taskonomy,cui2018large,pennington2014glove}, learning theoretic work~\citep{baxter2000model,maurer2006bounds,ben2010theory,rameshModelZooGrowing2022,tripuraneni2020provable,hanneke2020no,caruana1997multitask}, random matrix models~\citep{wei2022more}, neural tangent kernel models~\citep{malladi2022kernel} and information-theoretic analyses~\citep{jaakkola1999exploiting,achille2019task2vec,achilleDynamicsReachabilityLearning2019}. Broadly speaking, this work has focused on understanding the accuracy of a model on a new task when it is trained upon a related task, e.g., relationships between tasks are characterized using the excess risk of a hypothesis. Our methods also allow us to say things like ``task $P^1$ is far from $P^2$ as compared to $P^3$''. But they can go further. We can glean a global picture of the geometric structure in the space of tasks and quantify statements such as ``the divergence between $P^1$ and $P^2$ eventually is more than that of $P^1$ and $P^3$, but representations learned on these tasks are similar for 30\% of the way''. 

There is strong structure in typical inputs, e.g., recent work on understanding generalization~\citep{yang2021does,bartlett2020benign} as well as older work such as~\citet{simoncelliNaturalImageStatistics2001,fieldWhatGoalSensory1994,marrVisionComputationalInvestigation2010} has argued that visual data is effectively low-dimensional. Our works suggests that tasks also share a low-dimensional structure. Just like the effective low-dimensionality of inputs enables generalization on one task, effective low-dimensionality of the manifold of models trained on different tasks could perhaps explain generalization to new tasks.

\paragraph{Relationships between tasks in neuroscience}
Our results are conceptually close to those on organization and representation of semantic knowledge~\citep{mandler1993concept}. Such work has primarily used simple theoretical models, e.g., linear dynamics of~\citet{saxe2019mathematical} (who also use MDS). Our tools are very general and paint a similar picture of ontologies of complex tasks.
Concept formalization and specialization over age~\citep{vosniadou1992mental} also resembles our experiment in how fine-tuning models trained for longer periods changes the representation marginally. Our broad goals are similar to those of~\citet{sorscher2021geometry} but our techniques are very different.

\paragraph{Information Geometry} has a rich body of sophisticated ideas~\citep{amariinformationgeometryits2016}, but it has been difficult to wield it computationally, especially for high-dimensional models like deep networks. Our model in~\cref{eq:def:Pw} is a finite-dimensional probability distribution, in contrast to the standard object in information geometry which is an infinite-dimensional probability distribution defined over the entire domain. This enables us to compute embeddings of manifolds, geodesics, projections etc. We are not aware of similar constructions in the literature.

\paragraph{Visualizing training trajectories of deep networks}
InPCA is a variant of multi-dimensional scaling (MDS, see~\citet{cox2008multidimensional}), with the difference being that InPCA retains the negative eigenvalues which preserves pairwise distances~\citep{quinn2019visualizing}.
A large number of works have investigated trajectories of deep networks and the energy landscape during or after training using dimensionality reduction techniques~\citep{horoi2021visualizing,li2017visualizing,huang2020understanding}. \citet{gur2018gradient,antognini2018pca} studied the dimensionality of training trajectories. The key distinction here with respect to this body of work is that we study the prediction space, not the weight space. While the weight space has symmetries~\citep{freeman2016topology,garipov2018loss} and nontrivial dynamics~\citep{tanaka2021noether,chaudhari2017stochastic}, the prediction space, i.e,. $[0,1]^{N \times C} \ni \{p_w(c \mid x_i)\}$, completely characterizes the output of a probabilistic model. In comparison, the loss or the error which are typically used to reason about relationships between tasks, are coarse summaries of the predictions. Any two models, irrespective of their architecture, training methodology, or even the task that they were trained on, can be studied rigorously using our techniques.

\section*{Acknowledgments}
\label{s:ack}
RR, JM, RY and PC were supported by grants from the National Science Foundation (IIS-2145164, CCF-2212519), the Office of Naval Research (N00014-22-1-2255), and cloud computing credits from Amazon Web Services. IG was supported by the National Science Foundation (DMREF-89228, EFRI-1935252) and Eric and Wendy Schmidt AI in Science Postdoctoral Fellowship. HKT was supported by the National Institutes of Health (1R01NS116595-01). JPS was supported by the National Science Foundation (DMR-1719490), MKT was supported by the National Science Foundation (DMR-1753357). The authors would like to acknowledge Jay Spendlove for helpful comments on this material.

\clearpage
\bibliography{bib/contrastive,bib/pratik,bib/model_zoo,bib/reference_prior}
\bibliographystyle{icml2023}

\clearpage

\renewcommand\thefigure{A\arabic{figure}}
\renewcommand\thetable{A\arabic{table}}
\setcounter{figure}{0}
\setcounter{table}{0}

\appendix
\onecolumn

\section{Details of the experimental setup}
\label{s:app:setup}

\paragraph{Data}\mbox{}\\
We performed experiments using two datasets.
\begin{enumerate}[nosep]
    \item CIFAR10~\citep{krizhevsky2009learning} has 10 classes (airplane, automobile, bird, cat, deer, dog, frog, horse, ship, truck) with RGB images of size 32$\times$32, and
    \item ImageNet~\citep{dengImagenetLargescaleHierarchical2009} has 1000 classes each with about 1000 RGB images of size 224$\times$224.
\end{enumerate}

ImageNet classes are derived from the leaves of the Wordnet hierarchy~\citep{miller1998wordnet} which is visualized by~\citet{imagenet_bostock}. We use this hierarchy to create tasks using different subsets of ImageNet; We use all classes under a node to create a task. The tasks that we consider are: Dogs, Vertebrates, Invertebrates, Instrumentality, Reptile and Birds. We also consider a task with 333 randomly selected classes and unlike other tasks, it spans many different phyla of ImageNet.

\paragraph{Architectures}
We use a Wide-Resnet~\citep{zagoruyko2016wide} architecture for supervised learning experiments on CIFAR-10 (WRN-16-4 with depth 16 and widening factor of 4) and a Resnet-18~\citep{he2016deep} to train a model using SimCLR. All experiments on ImageNet use the Resnet-50 architecture.

All convolutional layers are initialized using the Kaiming-Normal initialization. For the Wide-Resnet, the final pooling layer is replaced with an adaptive pooling layer in order to handle input images of different sizes.

We make three modifications to these architectures.
\begin{enumerate} \itemsep 0em
\item  We remove the bias from the final classification layer; this helps keep the logits of the different tasks on a similar scale. 
\item In the experiments for Result 3 (episodic meta-learning) and Result 6 (fine-tuning), we replace batch normalization with layer norm in the Wide-Resnet. This is because we found in preliminary experiments that batch-normalization parameters make training meta-learning models very sensitive to choices of hyper-parameters (e.g., the support or query shot), and that the learned representations of new tasks were quite different in terms of their predictions (and thereby the Bhattacharyya distance) but all the difference was coming from modifications to the BN parameters.
\item In the Resnet-50, we replace the pooling layers with BlurPool~\citep{zhang2019making}. The bias parameter in batch normalization is set to zero with the affine scaling term set to one. 
\end{enumerate}
 
\paragraph{Training procedure}
All models are trained in mixed-precision (32-bit weights, 16-bit gradients) using stochastic gradient descent (SGD) with Nesterov's acceleration with momentum coefficient set to 0.9 and cosine annealing of the learning rate schedule. Batch-normalization parameters are excluded from weight decay.

CIFAR10 datasets use padding (4 pixels) with random cropping to an image of size 28$\times$28 or 32$\times$32 respectively for data augmentation. CIFAR10 images additionally have random left/right flips for data augmentation. Images are finally normalized to have mean 0.5 and standard deviation 0.25. 

Supervised learning models (including fine-tuning) for CIFAR10 are trained for 100 epochs with a batch-size of 64 and weight decay of $10^{-5}$ using the Wide-Resnet. 

Episodic meta-learners are trained using a Wide-Resnet and with the prototypical loss~\citep{snell2017prototypical}. For the 2-way meta-learner, each episode contains 20 query samples and 10 support samples. For the 5-way meta-learner, each episode contains 50 query samples and 10 support samples. We found (Result 4) to hold across different choices of these hyper-parameters in small-scale experiments. Models are trained for around 750 epochs and the episodic learner is about 5 times slower to train with respect to wall-clock time.

We train models using SimCLR on CIFAR10 and on tasks created from CIFAR10. For the augmentations, we use random horizontal flips, random grayscale, random resized crop and color jitter. Models are trained for 200 epochs for 2-way classification problems and for 500 epochs when trained on the entirety of CIFAR10 with the Adam optimizer and an initial learning rate of 0.001. 

\subsection{Experiments on ImageNet}

We make use of FFCV~\citep{leclerc2022ffcv}. which is a data-loading library that replaces the pytorch Dataloader. FFCV reduces the training time on ImageNet to a few hours, which allows us to train 100s of models on ImageNet, or on tasks created from it. Our implementation of ImageNet training builds on the FFCV repository \footnote{\href{https://github.com/libffcv/ffcv-imagenet/tree/main}{https://github.com/libffcv/ffcv-imagenet/tree/main}}.

ImageNet models are trained for 40 epochs with progressive resizing -- the image size is increased from 160 to 224 between the epochs 29 and 34. Models are trained on 4 GPUs with a batch-size of 512. The training uses two types of augmentations -- random-resized crop and random horizontal flips. Additionally, we use label smoothing with the smoothing parameter set to 0.1.




\subsection{Implementing InPCA in very high dimensions}
\label{app:inpca_extremely_high_dimensions}
We calculate an InPCA embedding of models along multiple trajectories, e.g., a typical experiment has about 25 trajectories (multiple random seeds, tasks, or representation learning methods) and about 50 models (checkpoints) along each trajectory. Each model is a very high-dimensional object (with dimensionality $N C$ where $N  \sim  10^5$ and $C  \sim 10$-$10^3$). Even if the matrix $D$ in~\cref{eq:w} is relatively manageable with $n \sim 1250$, each entry of $D$ is $\dB(P_u, P_v)$ and therefore requires $ \sim  10^8$ operations to compute. Implementing InPCA---or even PCA---for such large matrices requires a large amount of RAM. We reduced the severity of this issue to an extent using Numpy's memmap functionality \href{https://numpy.org/doc/stable/reference/generated/numpy.memmap.html}{https://numpy.org/doc/stable/reference/generated/numpy.memmap.html}. Also note that calculating only the top few eigenvectors of~\cref{eq:w} suffices to visualize the models, we do not need to calculate all.

The formula~\cref{eq:dB} is an effective summary of the discrepancies between how the predictions made by two probabilistic models differ; even small differences in two models, e.g., even if both $P_u$ and $P_v$ make mistakes on exactly the same input samples, if $p_u^n(c)$ is slightly different than $p_v^n(c)$ for even one of $n$ or $c$, the divergence is non-zero. InPCA is capable of capturing the differences between two such models~\cref{eq:w}. However, when the number of classes is extremely large, the number of terms in the summation is prohibitively large and analyzing the discrepancies or calculating the embedding becomes rather difficult.

We also developed a method to work around this issue. We can use a random stochastic matrix (whose columns sum up to 1) to project the outputs for each sample $\{p_u^n(c)\}_{c=1,\ldots,C}$ into a smaller space before calculating~\cref{eq:dB}. This amounts to pretending as if the model predicts not the actual classes but a random linear combination of the classes (even if the model is trained on the actual classes). This is a practical trick that is necessary only when we are embedding a very large number of very high-dimensional probabilistic models. We checked in our Imagenet experiments that using this trick gives the same embeddings.

In this paper, we did not need to use this projection trick. However, we found that this tricks makes it computationally faster to compute the embeddings and we have seen it to work well in practice. We have shared the code for this procedure, since it allows other people to reproduce the results using fewer computational resources.

\section{Additional Result}
\label{s:app:results}

\begin{figure}[H]
    \centering
    \includegraphics[width=0.95\textwidth]{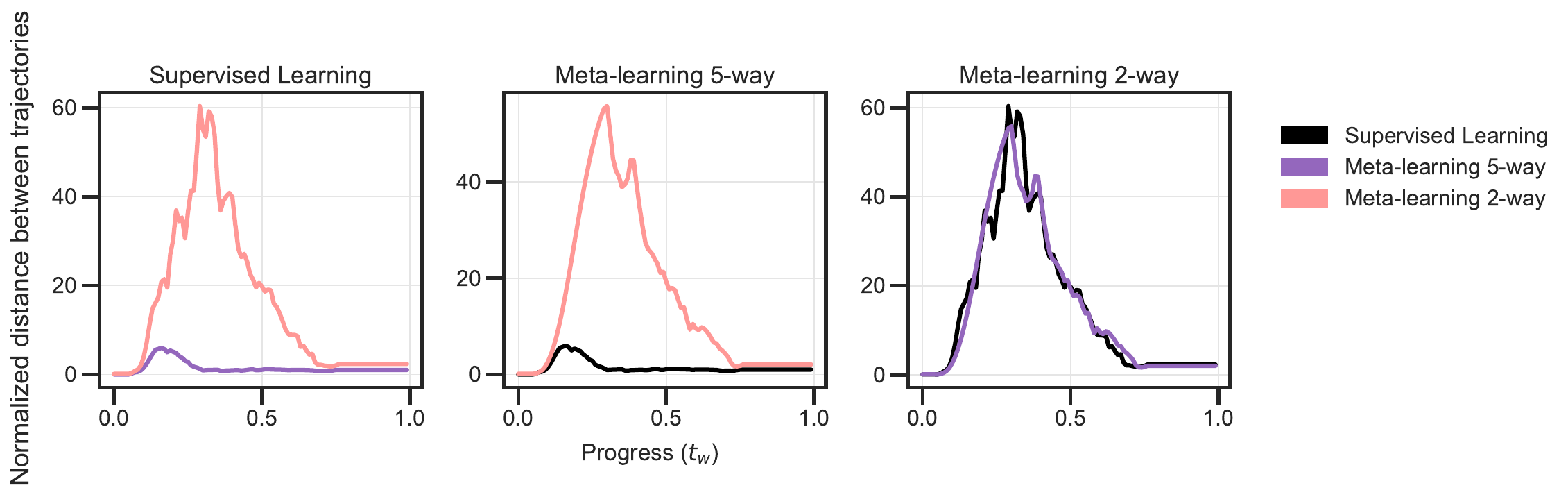}
    \caption{\textbf{Distance between trajectories of supervised and meta-learning at different values of progress.} Distances between the average trajectories of different algorithms (e.g., 2-way episodic learning and supervised learning, and 5-way episodic learning and supervised learning in the leftmost panel) are normalized by the average of the radii of the tubes corresponding to each trajectory. We find that trajectories of 2-way meta-learning deviate significantly from those of supervised learning for a large fraction of the trajectory. On the other hand, 5-way meta-learning is similar to the supervised learning trajectory for almost the entirety of the trajectory.}
    \label{fig:app:meta_dist}
\end{figure}

\begin{figure}
\centering
    \begin{subfigure}[c]{0.88\linewidth}
        \centering
        \includegraphics[width=\textwidth]{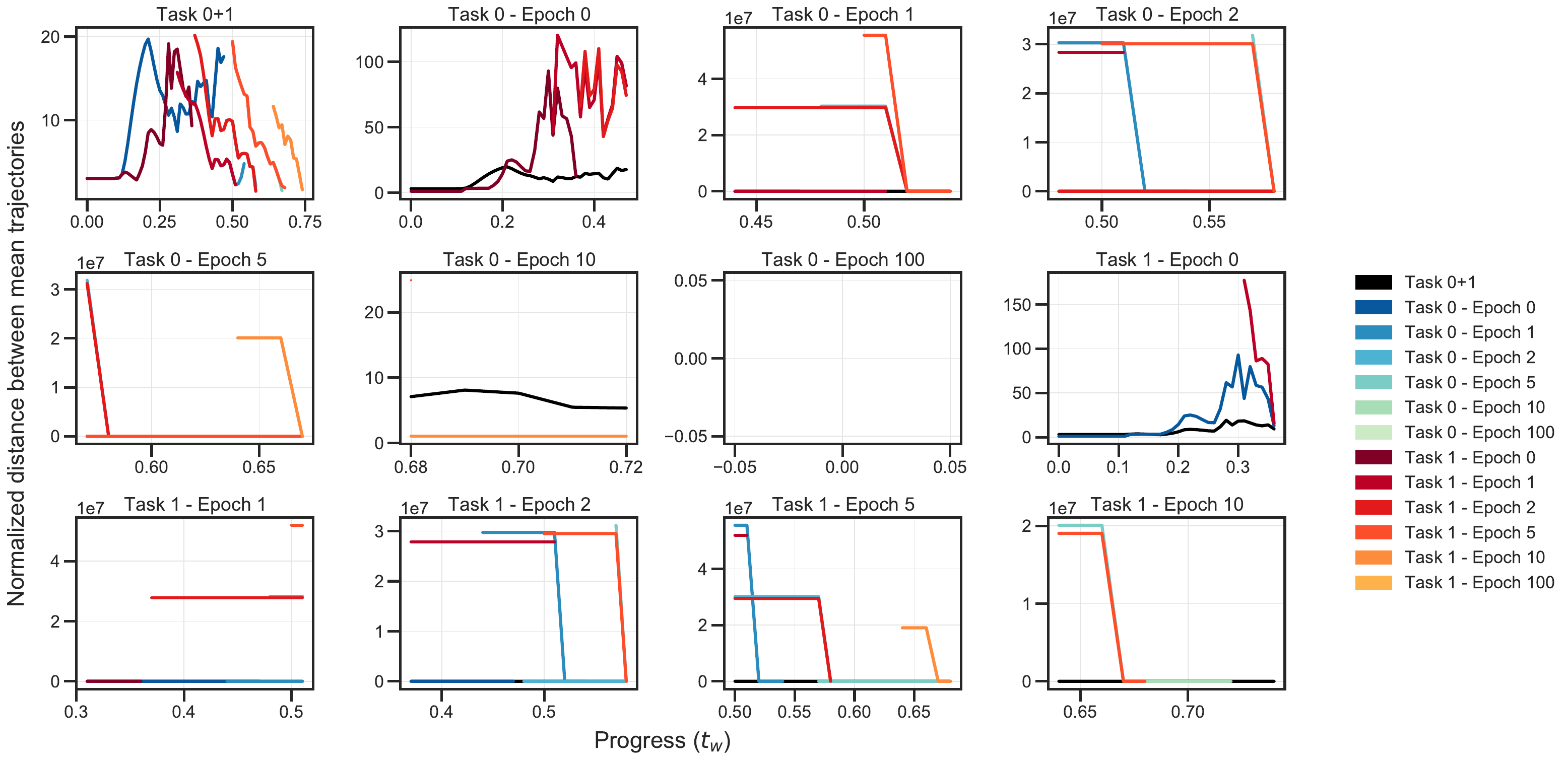}
        \caption{}\label{fig:app:tune_dist}
    \end{subfigure}
    \begin{subfigure}[c]{0.88\linewidth}
        \centering
        \includegraphics[width=\textwidth]{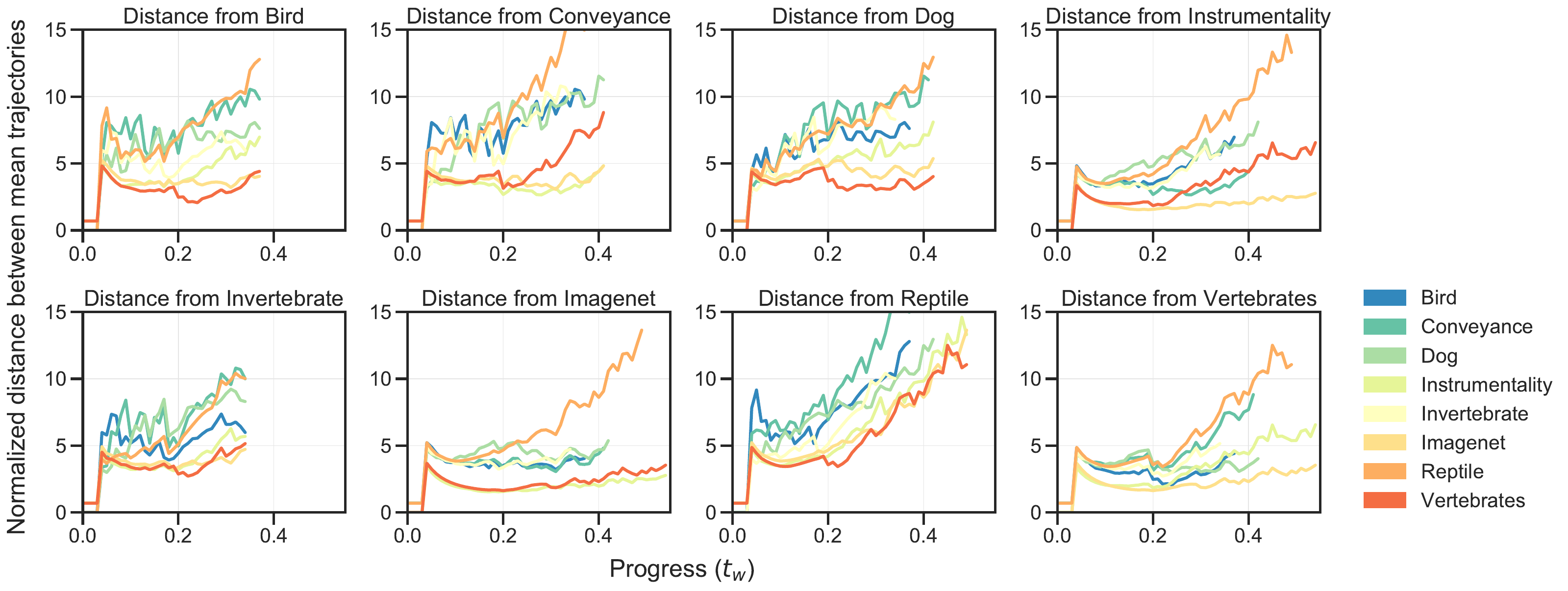}
        \caption{}\label{fig:app:ssl_dist_all}
    \end{subfigure}
    \begin{subfigure}[c]{0.88\linewidth}
        \centering
        \includegraphics[width=\textwidth]{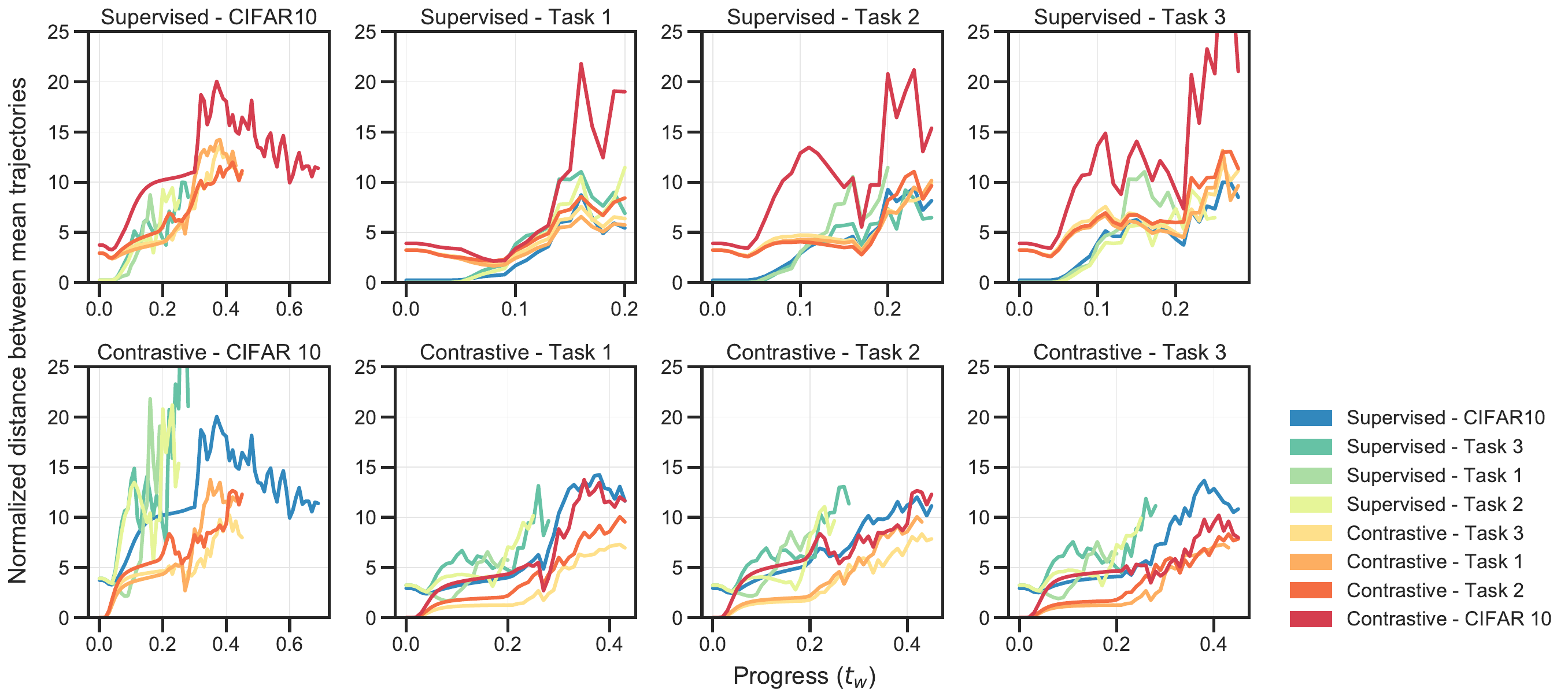}
        \caption{}\label{fig:app:ssl_dist}
    \end{subfigure}

\caption{This figure shows the extended version of the distances between trajectories of probabilistic models; two of them are identical to the ones in~\cref{fig:fine_tuning_panel}b and \cref{fig:ssl_panel}b.
}
\label{fig:app:trajectory_distances}
\end{figure}

\section{Bhattacharyya Distance}
\label{s:bhatt}

We provide additional details regarding~\cref{eq:dB}. Let $\bar y = (y_1, \cdots y_N)$, denote the labels assigned to each of the $N$ samples. Since there are $C$ classes in total, $\yvec$ can take a total of $C^N$ different values denoted by the set $Y^N$. Given, two models $P_u$ and $P_v$, the Bhattacharyya distance averaged over the samples is
\beqs{
    \aed{
        \dB(P_u, &P_v)
        := - N^{-1} \log (\sum_{\yvec \in Y^N} \sqrt{P_u(\bar y) P_v(\bar y)}) \\
        &= - N^{-1} \log (\sum_{\yvec \in Y^N} \prod_{n=1}^N\sqrt{p_u(y_n)\ p_v(y_n)}) \\
        &= - N^{-1}\log \left(\sum_{y_1=1}^C \sum_{y_2=1}^C \cdots \sum_{y_N=1}^C \left( \prod_{n=1}^{N} \sqrt{p_u(y_n)\ p_v(y_n)} \right) \right)\\
       &= - N^{-1}\log \left(\prod_{i=1}^N \left(\sum_{y_i=1}^C \sqrt{p_u(y_i)\ p_v(y_i)} \right) \right)\\ 
      &= - N^{-1} \sum_{i=1}^N \log  \left(\sum_{y_i=1}^C \sqrt{p_u(y_i)\ p_v(y_i)} \right).
    }
    \label{eq:dB2}
}

Uncovering the structure of high-dimensional probabilistic models is difficult because most distances between probability distributions saturate with the dimensionality, e.g., the Hellinger distance which is a metric, is essentially equal to 2 in high-dimensions.~\citet[Figure 1]{quinn2019visualizing} illustrates how a high-dimensional model benefits from using the Bhattacharyya distance compared to using the Hellinger distance in uncovering the intrinsic structure of the manifold. We believe that the logarithm in the Bhattacharyya distance keeps it well-behaved. We actually know of one other distance that gives meaningful results and that is the symmetric KL-divergence~\citep{teohVisualizingProbabilisticModels2020}, for the same reason: due to the logarithm. All analysis in our paper can therefore be done with the symmetric-KL divergence (which is also not a metric) and the results do look similar. 

There are a couple more reasons that motivated us to use the Bhattacharyya distance. First, the Bhattacharyya distance to the truth $P_*$ is equal to one half of the cross-entropy loss. Second, it is reassuring that both the Bhattacharyya distance locally gives the Fisher Information Matrix, which is positive semi-definite and therefore induces a local metric.

Bhattacharyya distance violates the triangle inequality and we speculate that this is necessary in order to uncover the low-dimensional structure in high-dimensional data. Understanding why it is important to violate the triangle inequality is a deep question and we do not know how to answer it yet. We do not use the Bhattacharyya distance itself to say things like “task A is close to task B” and as a result, the conclusions do not suffer from the violation of the triangle inequality. We only say things like “training on task A is equivalent to training for 80\% progress on task B, or ”training using contrastive learning is equivalent to training using supervised learning for 25\% of the progress“.

\section{Imprinting as an alternative to training the final layer}
\label{s:app:imprinting}

Consider a total of $C$ classes. We would like to find weights $\{w_c\}_{c=1}^C$ that maximize the log-probability of the samples, under the constraint that for all $c \in C$, the norm of the weights $||w_c||$ is 1. Let  $\varphi(x)$ denote a internal representation of sample $x$. The log-probability 
\beq{
\label{eq:imprint_prob}
    \sum_{x: y_{x}=c} \log p(y=c \mid x) 
    = \sum_{x: y_{x}=c} w_c \cdot \varphi(x) 
    - \sum_{x: y_{x}=c} \log \left( \sum_{j=c}^C \exp \left( w_c \cdot \varphi(x) \right) \right),
}
is proportional to the inner-product $w_c \cdot \sum_{x: y_x=c} \varphi(x)$. Maximizing just this term under the norm constraint, we get the imprinted weights $\sum_i \phi(x_i^c) / ||\sum_{i} \phi(x_i^c)||$ as the solution. Deriving an analytical expression for the optimal value of $\{w_c\}_{i=1}^{n_c}$ is difficult and hence we use the imprinted weights as an approximate solution. In our experiments, we found that the imprinted weights achieve an accuracy close to the optimal weights while being significantly easier to compute.



\section{Invariant transformations of the internal representation}

The internal representations are invariant to orthogonal transformations provided that we use imprinting to define a probabilistic model. This is because the internal representations define the same probabilistic model ever after an orthogonal transformation. Consider two internal representation $\phi$ and $U \cdot \phi$ where $U$ is an orthogonal matrix. We note that the probabilistic model for $U \cdot \phi$ after imprinting is
\begin{align*}
    \log p_2(y=c \mid x_i) 
    &=  \frac{U \cdot \sum_{y_x=c} \phi(x)}{||U \cdot \sum_{y_x=c} \phi(x)||} \cdot (U \cdot \phi(x_i)) 
    - \log \left( \sum_{c=1}^C \exp \left( 
         \frac{U \cdot \sum_{y_x=c} \phi(x)}{||U \cdot \sum_{y_{x}=c} \phi(x)||} \cdot (U \cdot \phi(x_i)) \right) \right) \\
    &=  \frac{\sum_{y_x=c} \phi(x)}{||\sum_{y_x=c} \phi(x)||} \cdot \phi(x_i)
    - \log \left( \sum_{c=1}^C \exp \left( 
         \frac{\sum_{y_x=c} \phi(x)}{||\sum_{y_{x}=c} \phi(x)||} \cdot \phi(x_i) \right) \right). \\
\end{align*}
The probabilistic model for the representation $U \cdot \phi$ is identical to the probabilistic model for representation $\phi$ since norms and angles are preserved under orthogonal transformations. Hence the Bhattacharyya distance between $\phi$ and $U \cdot \phi$ is zero.  

The imprinting procedure can be thought of as removing information from the representation that is not relevant to prediction on a task. While this is true for all datasets in general, there could exist some additional structure in the data that results in more invariances (e.g., more than invariances to orthogonal transformations $O(n)$).

\section{Measuring goodness-of-fit of an InPCA embedding using explained stress}
\label{s:explained_stress}

We would like to measure if a $k$-dimensional sub-space accurately preserves the true distances.
For this purpose, we define a quantity called the ``explained stress'' that estimates the fraction of pairwise distances in the original space that are preserved in the $k$-dimensional embedding. This is analogous to the explained variance in principal component analysis (PCA); but explained variance is a measure of the how well the original points are preserved in the embedding whereas explained stress approximates how well pairwise Bhattacharyya distances are preserved. If we consider the embedding to be given by first $k$ eigen-vectors, then the explained stress ($\chi_k$) is
\beq{
    \aed{
    \chi_k
    &= 1 - \frac{\norm{W - \sum_{i=1}^k \S_{ii}\ U_i U_i^\top}_{\text{F}}}{\norm{W}_{\text{F}}}
    = 1 - \sqrt{\frac{\sum_{i=k+1}^m \S_{ii}^2}{\sum_{i=1}^m \S_{ii}^2}}.
    }
    \label{eq:app:stress}
}
Note that InPCA finds an embedding that \emph{exactly} maximizes $\chi_k$.

\section{Calculating mean trajectories}
\label{s:app:mean_traj}


We defined the distance between two trajectories to be $\dtraj(\tou_u, \ttu_v)$, i.e., the integral of the Bhattacharyya distance between the trajectories after mapping them to the same task and re-indexing them using the geodesic. Say we wish to compare a model trained on two tasks from CIFAR-10: Cats vs. Dogs and Airplane vs. Truck. We initialize multiple models for each of these two supervised learning problems (and we do so for every experiment in this paper) and train these 10 models. We can now calculate the mean trajectory of models on a task
\[
    \argmin_{\t_{\mu}^1} \f{1}{K} \sum_{k=1}^K \dtraj(\t_{u_k}^1, \t_{\mu}^1).
\]
This optimization problem is very challenging because the variable is a trajectory of probabilistic models in a high-dimensional space. Even if we were to split this minimization and do it independently across time, this is still difficult because the solution is the so-called Bhattacharyya centroid on the product manifold defined in~\cref{eq:def:Pw} and cannot be computed in closed form. See~\citep{nielsen2011burbea} for an iterative formula.
We therefore simply take the arithmetic mean of the probability distributions, i.e., 
\(
    P_{\mu(t)} = \f{1}{K} \sum_{k=1}^K P_{w_i(t)}.
\)
This is similar to ensembling. We use the radius of the tube around the mean trajectory, i.e.,
\[
    r_u = \max_{k} \dtraj(\t_{u_k}^1, \t_{\mu}^1)
\]
to normalize distances (more precisely, we normalize using the average of the radii of the two trajectories being compared). Note that this radius depends upon time (and is computed after mapping and reindexing the trajectories). If the distance between the means of two sets of trajectories is smaller than their individual average radii, then the tubes around the means intersect each other. In such cases, one can say that the representations learned (at that time point) are not distinguishable.
We next show all distances between reindexed points along the trajectories discussed in~\cref{fig:imagenet1,fig:ssl_panel,fig:fine_tuning_panel}. Note that each curve gives the integrands in~\cref{eq:dtraj}, not the integral.

\section{Frequently Asked Questions}
\label{s:faq}

\begin{enumerate}

\item \textbf{These results are all intuitive and inline with literature. It is intuitive that trajectories of trained networks explore a small part of the prediction space; it is intuitive that training on one task makes progress on another task; it is intuitive that episodic meta-learning reaches the same solution as that of supervised learning, etc.}

Researchers working in the many sub-fields of machine learning discussed in this paper have various intuitions as to why their algorithms work. There are also numerous empirical results and theoretical models in these fields that are consistent with our findings. But this does not necessarily mean that we understand the underlying phenomena well. We must precisely quantify these intuitions and folklore results.

For example, it may be intuitive that episodic learning is ``similar'' to supervised learning because while the former uses a clustering loss over different sets of classes in each mini-batch, the latter uses a cross-entropy loss, which as we discussed in the main paper, will also lead to a good clustering of the features. Such intuition has been borne out in empirical results as well: fine-tuning-based few-shot learning methods have similar accuracy as that of episodic meta-learning-based ones. One may even argue intuitively that since these methods give a similar accuracy, the representations learned by these methods must be similar.

But intuition is a double-edged sword. For example, intuition also suggests that deep networks with millions of weights fitted on a non-convex energy landscape on seemingly dissimilar tasks are unlikely to learn similar representations. But since transfer learning is so remarkably effective, it is also intuitive that the representations of different tasks are similar; after all the network trains on the target task quickly after being pre-trained on the source task. Since networks are initialized randomly at different locations in the weight space, there is no reason to expect that the trajectory in prediction space will be low-dimensional. And yet, one might reason that since all these networks are trained on the same dataset, their training trajectories have to be similar to each other...

And this is why---to ground such intuition---we need precise quantitative studies. We have developed sophisticated techniques using information geometry to bear upon this problem. In some cases, we find surprising results---this opens up new avenues for theoretical and empirical investigation. In some cases our results may be consistent with the intuition of the practitioners---this lends credence to these techniques.

We are not aware of existing results in the literature which point out the phenomena identified in our paper.

\item \textbf{Do these findings translate to other tasks and other algorithms?}

We have used nine different tasks from ImageNet with very dissimilar classes and many sub-tasks of CIFAR-10, and five different representation learning methods (supervised learning, fine-tuning, episodic meta-learning, contrastive and semi-supervised learning).

The technical tools developed in this paper can indeed be used to study many other problem domains, algorithms and research questions. Each of these come with their own challenges, e.g., for tasks in natural language processing, although the probabilistic model is well-defined, the output space and the number of ``samples'' (words, sentences etc) is extremely large. As a rough estimate, for a typical book with $N \approx 10^5$ words, the output space has $C \approx 10^{20}$ if the model predicts the next 5 words. Therefore, although techniques developed here are well-defined, implementing and using them to answer new questions will require new research ideas. We hope that the research community will use these techniques to understand learned representations in the future. To aid this, we commit to releasing all code, data (model checkpoints, model embeddings), and interactive visualizations with the final version of the paper.

\item \textbf{How faithful are the visualizations of the trajectories in the prediction space?}

We are using a dimensionality reduction technique (InPCA) which has the specific property that if one uses all the eigenvalues (i.e., the dimensionality of the embedding space is equal to the number of probabilistic models being embedded) then the pairwise Bhattacharyya distances between points are preserved exactly. When we visualize the top three dimensions of the embedding, we only see a partial picture of the manifold and pairwise distances between points are no longer preserved. We can use the expression for explained stress~\cref{eq:explained_stress} and~\cref{eq:app:stress} to estimate how faithful our visualizations are. In our experience, the explained stress of the top three dimensions is always extremely high and this number is similar for all experiments in the paper. For example, the 2430 models of ImageNet in~\cref{fig:imagenet1} lie in a $10^7$-dimensional embedding space and yet the top 3 dimensions have an explained stress of about 80\%; the explained stress of the top 2430 dimensions of the embedding would be 100\%.

The visualizations in our paper are only used for providing intuition and interpretability; all findings in the paper are made quantitatively on the basis of the integral of the Bhattacharyya distances between points along the trajectories.

\item \textbf{Can you explain the time re-parameterization using the geodesic? It may not be a good normalization in all situations.}

Different models train at very different speeds, in particular at the beginning of training models move rapidly after each mini-batch update in the space of probability distributions (e.g., as measured by the Bhattacharyya distance). We can only record the trajectory at specific checkpoints (e.g., after each epoch or, at best, after each mini-batch). Our time re-parameterization technique allows us to interpolate two successive checkpoints using the geodesic that connects them. We discretize “progress” $t_w$ into 100 equidistant intervals and interpolate the entire trajectory at these 100 points by calculating the appropriate points on the geodesic between two successive checkpoints; this is described in the narrative after~\cref{eq:tw}. For initial parts of the trajectory, there are fewer checkpoints per unit progress because the network trains quickly. But the time re-parameterization still allows us to discretize the entire trajectory evenly. When we do the analysis using the distances between trajectories, this technique ensures equal importance to early and late times of the training process irrespective of how fast the network learns in these phases. Our techniques also allow us to quantitatively study the differences in how quickly different networks train, although this is not the focus of this paper.

\item \textbf{CIFAR10 is a very small dataset to apply semi, self-supervised learning.}

Self-supervised learning models have also been trained on CIFAR-10 by a number of papers~\citep{huang2021towards,zhong2022self,shen2022mix}. Contrastive learning has also been applied to smaller datasets in the context of unsupervised domain-adaptation~\citep{ruan2021optimal}. In our experiments, SimCLR trains to an accuracy of 87.36\% on CIFAR10 (after imprinting the final model), which is only about 7\% worse than a typical run of supervised learning in spite of not using any labels whatsoever.

For Result 5 on contrastive and semi-supervised learning, we chose CIFAR-10 for contrastive learning because it is incredibly expensive to train on ImageNet. \citet{zbontar2021barlow} report that their method requires 32 V100 GPUs over 124 hours. For all experiments in our paper, we train on 5 random seeds (to create trajectories from different initializations, interpolate them using geodesics, compute averages etc.). Running SimCLR on ImageNet to perform an analysis of its trajectories would be enormously expensive (about \$60,000 on AWS for 5 trajectories of one algorithm...).
\end{enumerate}

\end{document}